\documentclass[runningheads]{llncs}
\usepackage{graphicx}
\usepackage{tikz}
\usepackage{comment}
\usepackage{amsmath,amssymb}
\usepackage{color}
\usepackage{hyperref}

\usepackage{booktabs}
\usepackage{epsfig}
\usepackage{bbding}
\usepackage{multirow}
\usepackage{enumerate}
\usepackage{bm}
\usepackage{cite}
\usepackage{bbding}

\usepackage[accsupp]{axessibility}  % Improves PDF readability for those with disabilities.

\newcommand{\mypar}[1]{{\bf #1.}}

\usepackage[capitalize]{cleveref}
\crefname{section}{Sec.}{Secs.}
\Crefname{section}{Section}{Sections}
\Crefname{table}{Table}{Tables}
\crefname{table}{Tab.}{Tabs.}

\begin{document}
% \linenumbers
\pagestyle{headings}
\mainmatter
\def\ECCVSubNumber{4239}  % Insert your submission number here

\title{Skeleton-Parted Graph Scattering Networks for 3D Human Motion Prediction} % Replace with your title

%\begin{comment}
 \titlerunning{Skeleton-Parted Graph Scattering Networks (SPGSN)} 
 \authorrunning{M. Li et al.} 
 \author{Maosen Li\inst{1} \and
            Siheng Chen \inst{1,2} \Envelope \and
            Zijing Zhang \inst{3} \and
            Lingxi Xie \inst{4} \and  \\
            Qi Tian \inst{4} \and
            Ya Zhang \inst{1,2} \Envelope}
 \institute{Cooperative Medianet Innovation Center, Shanghai Jiao Tong University \and
                Shanghai AI laboratory \and 
                Zhejiang University \and 
                Huawei Cloud \& AI. \\ \email{\{maosen\_li, sihengc, ya\_zhang\}@sjtu.edu.cn, zijing\_zhang@163.com, 198808xc@gmail.com, tian.qi1@huawei.com} }
                
%\end{comment}

% \begin{comment}
\titlerunning{Skeleton-Parted Graph Scattering Networks (SPGSN)}
\authorrunning{M. Li et al.} 
 \author{Maosen Li\inst{1} \and
            Siheng Chen \inst{1,2} \Envelope \and
            Zijing Zhang \inst{3} \and
            Lingxi Xie \inst{4} \and  \\
            Qi Tian \inst{4} \and
            Ya Zhang \inst{1,2} \Envelope}
 \institute{Cooperative Medianet Innovation Center, Shanghai Jiao Tong University \and
                Shanghai AI laboratory \and 
                Zhejiang University \and 
                Huawei Cloud \& AI. \\ \email{\{maosen\_li, sihengc, ya\_zhang\}@sjtu.edu.cn, zijing\_zhang@163.com, 198808xc@gmail.com, tian.qi1@huawei.com} }
% \end{comment}
%******************
\maketitle

% \vspace{-5mm}
\begin{abstract}
Graph convolutional network based methods that model the body-joints' relations, have recently shown great promise in 3D skeleton-based human motion prediction. However, these methods have two critical issues: first, deep graph convolutions filter features within only limited graph spectrums, losing sufficient information in the full band; second, using a single graph to model the whole body underestimates the diverse patterns on various body-parts. To address the first issue, we propose adaptive graph scattering, which leverages multiple trainable band-pass graph filters to decompose pose features into richer graph spectrum bands. To address the second issue, body-parts are modeled separately to learn diverse dynamics, which enables finer feature extraction along the spatial dimensions. Integrating the above two designs, we propose a novel skeleton-parted graph scattering network (SPGSN). The cores of the model are cascaded multi-part graph scattering blocks (MPGSBs), building adaptive graph scattering on diverse body-parts, as well as fusing the decomposed features based on the inferred spectrum importance and body-part interactions. Extensive experiments have shown that SPGSN outperforms state-of-the-art methods by remarkable margins of $13.8\%$, $9.3\%$ and $2.7\%$ in terms of 3D mean per joint position error (MPJPE) on Human3.6M, CMU Mocap and 3DPW datasets, respectively~\footnote[1]{The codes are available at \url{https://github.com/MediaBrain-SJTU/SPGSN}.}.
% \vspace{-2mm}
% \vspace{-3mm}
\end{abstract}

\section{Introduction}
% \vspace{-3mm}
\label{sec:One_intro}
3D skeleton-based human motion prediction has attracted increasing attention and shown broad applications, such as human-computer interaction~\cite{gui-2018-110272} and autonomous driving~\cite{SihengChen_IEEE_SP}.
Human motion prediction aims to generate the future human poses, in form of the 3D coordinates of a few key body joints, given the historical motions. Early attempts develop state models~\cite{NIPS2006_3078,icml2009_129,Lehrmann_2014_CVPR,NIPS2000_ca460332} to capture the shallow dynamics. In the deep learning era, more implicit patterns are learned. For example, some recurrent-network-based methods~\cite{Fragkiadaki_2015_ICCV,Walker_2017_ICCV,Martinez_2017_CVPR,Gui_2018_ECCV} aggregate the states and predict poses frame-by-frame; some feed-forward models~\cite{Li_2018_CVPR,guo2019human} directly output the predictions without state accumulation.

\begin{figure}[t]
    \centering
    \includegraphics[width=0.9\columnwidth]{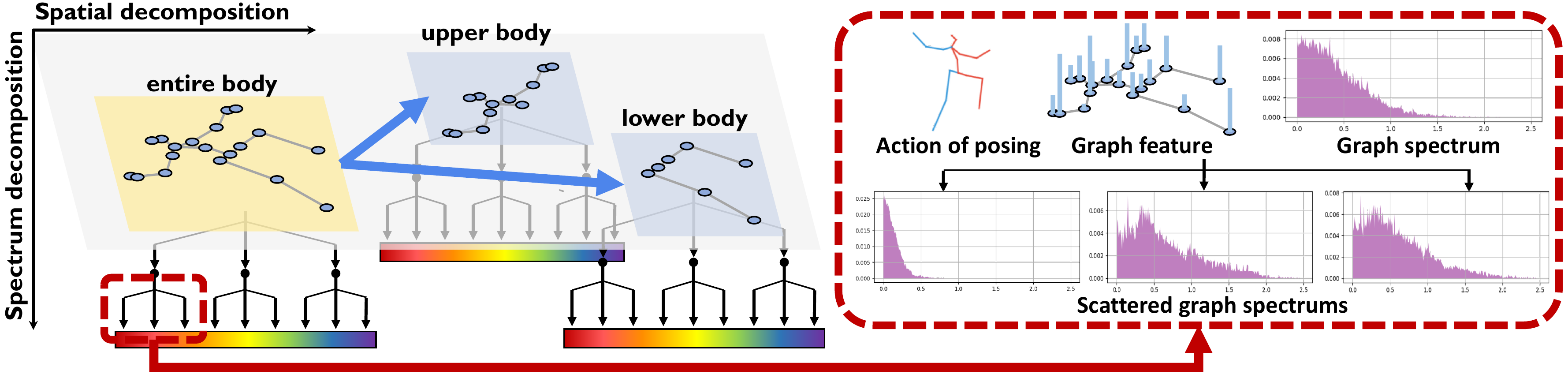}
    % \vspace{-2mm}
    \caption{\small Feature decomposition along the spatial and spectrum domains. For example, we separate the body into the upper and lower bodies, each of which uses a three-branch graph scattering tree with band-pass filtering to exploit rich graph spectrums.}
    \label{fig:jewel}
    % \vspace{-5mm}
\end{figure}

Recently, numerous graph-convolution-based models~\cite{Mao_2019_ICCV,Li_2020_CVPR,li2021multiscale,Cui_2020_CVPR,Mao_2020_ECCV,Dang_2021_ICCV,Liu_2021_ICCV,Sofianos_2021_ICCV} have achieved remarkable success in motion prediction by explicitly modeling the inherent body relations and extracting spatio-temporal features~\cite{kipf2018neural,Qi_2018_ECCV,Wang_2020_CVPR}.  However, further development of graph-based methods encounters two critical issues.  {First, as long as the graph structure is given, standard graph convolution just filters the features within limited graph spectrum but cannot significantly preserve much richer bands (e.g., smoothness and difference on the graphs) at the same time. However, the pattern learning of human motions needs not only to capture the similarity or consistency of body-joints under the spatial constraints (low-frequency), but also to enhance the difference for diverse representation learning (high-frequency). For example, when the graph edge weights are purely positive or negative, deep graph convolution tends to average the distinct body-joints to be similar but ignores their specific characteristics.}
% and results in information loss. {For example, when the graph edge weights are purely positive or negative, graph convolution only emphasizes the smoothness of the signal, and perform low-pass filtering by averaging the node features; when edge weights have alternating positive and negative distributions, graph convolution focuses the node difference. In other words, as long as the graph topology is given, the simple graph convolution only describes the information in a specific frequency band but cannot significantly preserve the much rich spectrums (e.g., smoothness and difference) at the same time.}
Second, existing methods usually use a single graph to model the whole body~\cite{Mao_2019_ICCV,Mao_2020_ECCV}, which  underestimates diverse movement patterns in different body-parts~\cite{guo2019human}. For example, the upper and lower bodies have 
distinct motions, calling for using different graphs to represent them separately.

To address the first issue, we propose the adaptive graph scattering technique, which leverages multiple trainable band-pass graph filters arranged in a tree structure to decompose input features into various graph bands. With the mathematically designed band-specific filters, adaptive filter coefficients and feature transform layers, it preserves information from large graph spectrum. To address the second issue, we decompose a body into multiple body-parts, where comprehensive dynamics could be extracted. Therefore, our method achieves finer feature extraction along both graph spectrum and spatial dimensions. Fig.~\ref{fig:jewel} sketches both the spatial and spectrum decomposition. As an example, we show three body-parts: the upper body, the lower body and the entire body. To understand the graph scattering, we show the output graph spectrums on different bands after the corresponding filtering processes.

\begin{figure*}[t]
    \centering
    \includegraphics[width=1\textwidth]{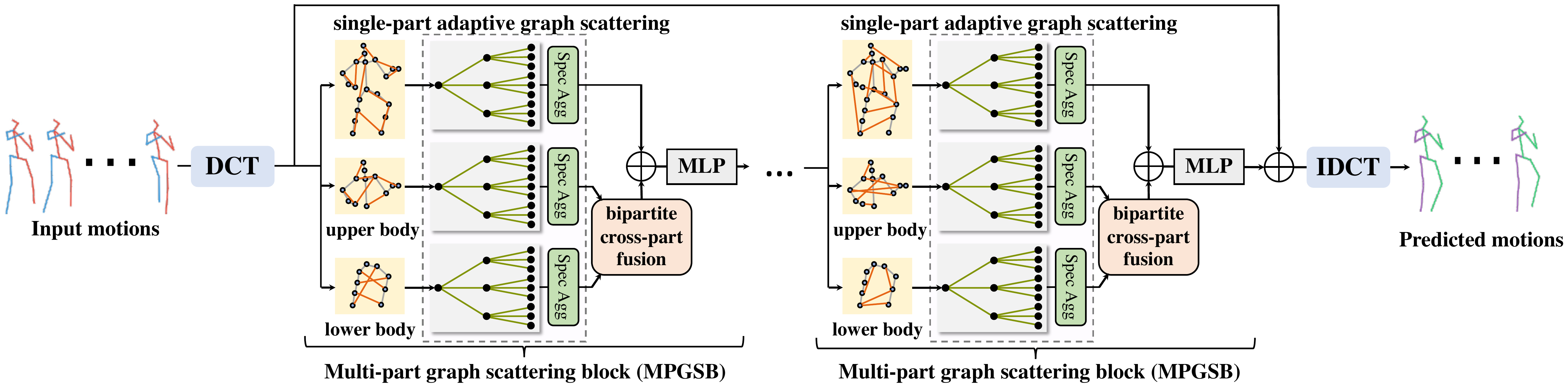}
    %\vspace{-4mm}
    \caption{\small Architecture of the SPGSN. The SPGSN first applies discrete cosine transform (DCT) to convert the body-joint positions along time to the frequency domain. Then, cascaded multi-part graph scattering blocks (MPGSBs) are built for deep feature extraction. 
    % Each MPGSB takes adaptive graph scattering on separated body-parts and fuse them into hybrid features.
    Finally, we build a skip-connection between input and output features and use inverse DCT (IDCT) to recover the temporal information.}
    \label{fig:architectures}
    % \vspace{-5mm}
\end{figure*}

Integrating the above two designs, we propose a novel \emph{skeleton-parted graph scattering network} (SPGSN). The core of SPGSN is the~\emph{multi-part graph scattering block} (MPGSB), consisting of two key modules: the single-part adaptive graph scattering, which uses multi-layer graph scatterings to extract spectrum features for each body-part, and bipartite cross-part fusion, which fuses body-part features based on part interactions. The SPGSN consists of multiple MPGSBs in a sequence; see Fig.~\ref{fig:architectures}. Taking the 3D motions as inputs, SPGSN first converts the feature along temporal dimension by discrete cosine transform (DCT) to obtain a more compact representation, which removes the complexity of temporal modeling~\cite{Mao_2019_ICCV,Mao_2020_ECCV}. Followed by the network pipeline, an inverse DCT recovers the responses to the temporal domain. A cross-model skip-connection is built to learn the residual DCT coefficients for stable prediction. 

Extensive experiments are conducted for both short-term and long-term motion prediction on large-scale datasets, i.e., Human3.6M~\cite{ionescu2013human3}, CMU Mocap\footnote{ http://mocap.cs.cmu.edu/} and 3DPW~\cite{Marcard_2018_ECCV}. Our SPGSN significantly outperforms state-of-the-art methods in terms of mean per joint position error (MPJPE). The main contributions of our work are summarized here:

$\bullet$ We propose the skeleton-parted graph scattering networks (SPGSN) to promote finer feature extraction along both graph spectrum and spatial dimensions, resulting in more comprehensive feature extraction in large graph spectrum and spatially diverse dynamics learning to improve prediction.

$\bullet$ In SPGSN, we develop the multi-part graph scattering block (MPGSB), which contains single-part adaptive graph scattering and cross-part bipartite fusion to learn rich spectral representation and aggregate diverse part-based features for effective dynamics learning.

$\bullet$ We conduct experiments to verify that our SPGSN significantly outperforms existing works by $13.8\%$, $9.3\%$ and $2.7\%$ in terms of MPJPE for motion prediction on Human3.6M, CMU Mocap and 3DPW datasets, respectively.

\section{Related Works}
% \vspace{-1mm}
\label{sec:One_rw}
\subsection{Human Motion Prediction}
% \vspace{-1mm}
For human motion prediction, early methods are developed based on state models~\cite{NIPS2006_3078,icml2009_129,Lehrmann_2014_CVPR,NIPS2000_ca460332}. Recently, some recurrent-network-based models consider the sequential motion states. ERD~\cite{Fragkiadaki_2015_ICCV} and Pose-VAE~\cite{Walker_2017_ICCV} build encoder-decoder in recurrent forms. Structural-RNN \cite{Jain_2016_CVPR} transfers information between body-parts recurrently. Res-sup~\cite{Martinez_2017_CVPR}, AGED~\cite{Gui_2018_ECCV} and TPRNN~\cite{Li_2020_CVPR} model the pose displacements in RNN models. Besides, some feed-forward networks use spatial convolutions to directly predict the whole sequences without state accumulation~\cite{Li_2018_CVPR,guo2019human}. Furthermore, considering an articulated pose, some methods exploit the correlations between body-components~\cite{Mao_2019_ICCV,Cui_2020_CVPR,Sofianos_2021_ICCV,Mao_2020_ECCV,Cai_ECCV_2020}. DMGNN~\cite{Li_2020_CVPR} and MSR-GCN~\cite{Dang_2021_ICCV} build multiscale body graphs to capture local-global features. TrajCues~\cite{Liu_2021_ICCV} expands motion measurements to improve feature learning.
Compared to previous models, our method leverages rich band-pass filters to preserve both smoothness and diversity of body joints and achieve more precise prediction.

% \vspace{-2mm}
\subsection{Graph Representation Learning}
% \vspace{-1mm}
Graphs explicitly depict the structural format~\cite{AAAI1817135,8779725} of numerous data, such as social networks~\cite{tabassum2018social,NEURIPS2020_a26398dc}, human poses and behaviors~\cite{Qi_2018_ECCV,Wang_2020_CVPR,8842613,AAAI1817135,xu2021invariant,Shi_2019_CVPR,Hu_ICME_2019,9062552,9067002,9102429,Fan_2019_ICCV,Lu_ECCV_2020}, and dynamic systems~\cite{kipf2018neural,xu2022groupnet,huang2019stgat,kosaraju2019social,li2020evolvegraph,Hu_2020_CVPR}. As effective methods of graph learning, some studies of the graph neural networks (GNNs) are developed to perform signal filtering based on the graph Laplacian eigen-decomposition~\cite{8745502,Bruna2014ICLR,NIPS2016_6081,kipf_iclr2017} or to aggregate vertex information~\cite{NIPS2017_6703,pmlr-v70-Niepert16,velickovic2018graph,Li2016ICLR,ICML_2016_Dai}.
Recently, graph scattering transform (GST) and related models are developed, promoting to capture rich graph spectrum with large bandwidth  ~\cite{Ioannidis2020Pruned,Gao_2019_ICML_GS,NEURIPS2020_a6b964c0,pan2021spatiotemporal}. GSTs generalize the image-based scattering transforms~\cite{6522407,Sifre_2013_CVPR,6822556,NEURIPS2019_3ce3bd7d}, combining various graph signal filters with theoretically justified designs in terms of spectrum properties. 
% \cite{NEURIPS2019_3ce3bd7d} proves the stability for GSTs. 
\cite{gama2018diffusion,ZOU20201046} develop diffusion wavelets. \cite{NEURIPS2020_a6b964c0,9414557} integrate designed scattering filters and parameterized feature learners. \cite{pan2021spatiotemporal} expands GSTs on the spatio-temporal domain. 
In this work, we employ mathematical prior to initialize an adaptive graph scattering with trainable band-pass filters, filter coefficients and feature mapping.

\section{Skeleton-Parted Graph Scattering Network}
\label{sec:One_method}
%\vspace{-2mm}

\subsection{Problem Formulation}
\label{sec:Two_preliminary}
%\vspace{-1mm}

Skeleton-based motion prediction aims to generate the future poses given the historical ones.  Mathematically, let  $\mathbf{X}^{(t)}\in\mathbb{R}^{M \times 3}$  be a pose carrying the 3D coordinates of $M$ body joints at time $t$,   ${\mathbb{X}} = [\mathbf{X}^{(1)},\dots,\mathbf{X}^{(T)}] \in \mathbb{R}^{T \times M \times 3}$  be a three-mode tensor that concatenates moving poses within $T$ timestamps. In motion prediction, let ${\mathbb{X}^{-}} = [\mathbf{X}^{(-T+1)},\dots,\mathbf{X}^{(0)}] {\in \mathbb{R}^{ T \times M \times 3}}$ represent $T$ historical poses, ${\mathbb{X}^{+}} = [\mathbf{X}^{(1)},\dots,\mathbf{X}^{(\Delta T)}] {\in \mathbb{R}^{\Delta T \times M \times  3}}$ represent $\Delta T$ future poses. We aim to propose a predictor $\mathcal{F}_{\rm pred}(\cdot)$ to predict the future motions $\widehat{\mathbb{X}}^{+}=\mathcal{F}_{\rm pred}(\mathbb{X}^{-})$ to approximate the ground-truth $\mathbb{X}^{+}$. 

% In this work, we consider the human-bodies as distinct body-parts and build independent graphs on them to model their internal relations. We set three body-parts: the whole body, the upper body and the lower body, which have $M$, $M_{\uparrow}$ and $M_{\downarrow}$ joints, respectively. Note that we have $M = M_{\uparrow} + M_{\downarrow}$. On the three separated bodies, we define the corresponding body-part graphs as $G(\mathcal{V},\mathbf{A})$, $G_{\uparrow}(\mathcal{V}_{\uparrow},\mathbf{A}_{\uparrow})$ and $G_{\downarrow}(\mathcal{V}_{\downarrow},\mathbf{A}_{\downarrow})$, where $\mathcal{V}$, $\mathcal{V}_{\uparrow}$, $\mathcal{V}_{\downarrow}$ are the vertex sets of the three graphs; $\mathbf{A}\in\mathbb{R}^{M \times M}$, $\mathbf{A}_{\uparrow}\in\mathbb{R}^{M_{\uparrow} \times M_{\uparrow}}$ and $\mathbf{A}_{\downarrow}\in\mathbb{R}^{M_{\downarrow} \times M_{\downarrow}}$ are the corresponding adjacency matrices, respectively.

%\vspace{-2mm}
\subsection{Model Architecture}
\label{sec:Two_architecture}
% \vspace{-1mm}

Here we propose the model architecture and the operation pipeline of the \emph{Skeleton-Parted Graph Scattering Network} (SPGSN), which is sketched in  Fig.~\ref{fig:architectures}. 

Taking the historical motion tensor ${\mathbb{X}}^{-} = [\mathbf{X}^{(1)},\dots,\mathbf{X}^{(T)}]$ as the input, we first apply the discrete cosine transform (DCT) along the time axis to convert the temporal dynamics of motions into the frequency domain, leading to a compact representation that eliminates the complexity of extra temporal embedding to promote easy learning~\cite{Mao_2020_ECCV,Mao_2019_ICCV}. Mathematically, we reshape $\mathbb{X}^{-}$ into $\mathcal{X}^- \in \mathbb{R}^{T \times 3M}$ to consider all the joint coordinates at each timestamp independently as the basic units in the spatial domain; then we encode ${\bf X}^{-} = {\rm DCT}(\mathcal{X}^{-})\in\mathbb{R}^{M' \times C}$, where $M'=3M$, and $C$ denotes the number of DCT coefficients, also the feature dimension. In this way, although we triple the spatial scale, we compress the long sequence into a compact coefficient representation, resulting in a feature vector, and we do not need the additional sequential feature modeling. Compared to other frequency transformations, DCT fully preserves the temporal smoothness. The Fourier transform and wavelet transform usually introduce complex and multiscale responses, making the downstream modeling more complicated.

In the SPGSN, we develop a deep feed-forward architecture to learn the dynamics from the DCT-formed motion features ${\bf X}^{-}$. The network is constructed with cascaded \emph{multi-part graph scattering blocks} (MPGSBs) as the core components. All MPGSBs do not share parameters, and the input of the following MPGSB is the output of the last one. In each MPGSB, the input motion is first decomposed into different body-parts. For example, Fig.~\ref{fig:architectures} sketches the entire body, the upper and lower bodies, but different body separation strategies could be employed. There are trainable graphs on these body-parts. On each body-part, MPGSB takes a single-part adaptive graph scattering to preserve large-band spectrums of motion representation (see Sec. \ref{sec:Two_SPAGS}). On multiple body-parts, an bipartite cross-part fusion automatically performs body-part fusion based on the learned cross-part interaction for more coordinated motion estimation (see Sec. \ref{sec:Two_Fuse}). 
Moreover, we build skip connections across all the MPGSBs, thus we force the SPGSN to capture the feature displacements for stable prediction. At the output end, we apply inverse DCT to recover the temporal information. 

% \vspace{-2mm}
\section{Multi-Part Graph Scattering Block}
% \vspace{-1mm}
\label{sec:One_MPGSB}
Here we present the \emph{Multi-Part Graph Scattering Blocks} (MPGSBs). Each MPGSB contains two key modules, \textbf{1) single-part adaptive graph scattering} and \textbf{2) bipartite cross-part fusion}, to extract large graph spectrum from distinct body-parts and fuse body-parts into hybrid representation, respectively.

% \vspace{-2mm}
\subsection{Single-Part Adaptive Graph Scattering}
\label{sec:Two_SPAGS}
% \vspace{-1mm}
Given the determined graph topology, the information-aggregation-based graph filtering only captures features in limited spectrums, losing the rich frequency bands of the inputs. To address this problem, the single-part adaptive graph scattering learns sufficient large-band information on each body-part.
The single-part adaptive graph scattering contains two main operations: 1) \emph{adaptive graph scattering decomposition} and 2) \emph{adaptive graph spectrum aggregation}, which parses graph spectrum and aggregates the important bands, respectively.

\mypar{Adaptive graph scattering decomposition}
The adaptive graph scattering decomposition forms a tree-structure network with $L$ layers and exponentially increasing tree nodes. These tree nodes perform graph filtering on the motion corresponding to various graph spectral bands. 

We consider the first layer for example, and we could expand the design to any layers. Let the DCT-formed pose feature be ${\bf X}\in\mathbb{R}^{M'\times C}$, and the adaptive pose graph have adjacency matrix ${\bf A}\in\mathbb{R}^{M' \times M'}$. We utilize a series of band-pass graph filters: $\{h_{(k)}(\widetilde{\bf A})|k=0,1,\dots,K\}$, which are derived based the graph structure. Note that we have the normalized $\widetilde{\bf A} = 1/2 ({\bf I}+{\bf A}/{\|{\bf A}\|_F^2})$ to handles the amplitudes of the trainable elements. Given the filter bank, $\{ h_{(k)}(\widetilde{\bf A}) \}_{k=0}^{K}$, we obtain features $\{ {\bf H}_{(k)}\in\mathbb{R}^{M' \times C'} \}_{k=0}^{K}$ through
\begin{equation}
    \label{eq:GNN_with_filter}
    \setlength{\abovedisplayskip}{5pt}
    \setlength{\belowdisplayskip}{5pt}
    {\bf H}_{(k)}=
    \sigma(h_{(k)}(\widetilde{\bf A}){\bf X}{\bf W}_{(k)}),
\end{equation}
where ${\bf W}_{(k)}$ is the trainable weights corresponding to the $k$th filter, and the nonlinear $\sigma(\cdot)$ (e.g. Tanh) disperses the graph frequency representation~\cite{Ioannidis2020Pruned}. Note that $\widetilde{\bf A}$ is also a parameterized matrix automatically tuned during training to adapt to the  implicit interaction in motion data.

\begin{figure}[t]
    \centering
    \includegraphics[width=0.7\columnwidth]{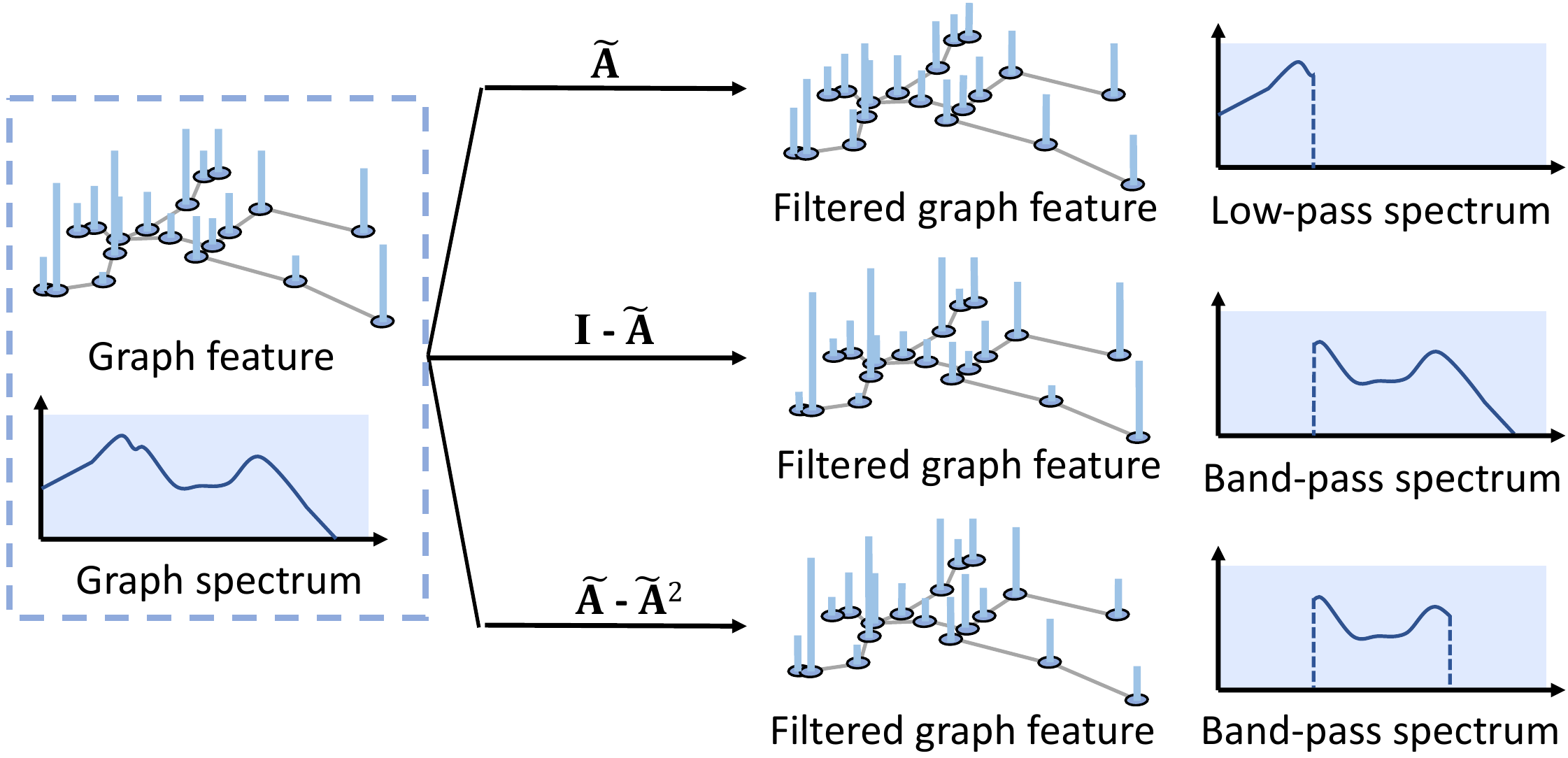}
    %\vspace{-2mm}
    \caption{\small Sketch of the graph filtering with graph filters. Note that we just show the initialized filters, and we apply trainable coefficients to achieve more flexible filtering.}
    \label{fig:scatterprinciple}
    %\vspace{-5mm}
\end{figure}

To ensure that various filters work on specific graph spectrums, we initialize $\{ h_{(k)}(\widetilde{\bf A}) \}_{k=0}^{K}$ by leveraging the mathematical priors to constrain their filtering bands. Furthermore, we apply trainable coefficients in $\{ h_{(k)}(\widetilde{\bf A}) \}_{k=0}^{K}$ to adaptively tune spectrum responses based on the predefined guidance; that is,
\begin{equation}
  \label{eq:wavelets}
  \setlength{\abovedisplayskip}{2mm}
  \setlength{\belowdisplayskip}{2mm}
  \begin{aligned}
    h_{(k)}(\widetilde{\bf A}) &= \alpha_{(0,0)}\widetilde{\bf A}, & k = 0; &\\
    h_{(k)}(\widetilde{\bf A}) &= \alpha_{(1,0)}{\bf I} + \alpha_{(1,1)}\widetilde{\bf A}, & k = 1; &\\
    h_{(k)}(\widetilde{\bf A}) &= \sum\nolimits_{j=1}^{k} \alpha_{(k,j)} \widetilde{\bf A}^{2^{j-1}}, &k =2,\dots,K,&
  \end{aligned}
\end{equation}
where $\alpha_{(k,j)}$ is the trainable coefficient. For $k=0$, we initialize $\alpha_{(0,0)}=1$; for $k>0$, we set $\alpha_{(k,k-1)}=1$, $\alpha_{(k,k)}=-1$ and any other $\alpha_{(k,j)}=0$. Notably, for $k>0$, we could approximately obtain a series of graph wavelets to emphasize different frequencies. For example, we initialize $h_3(\widetilde{\bf A})=0\widetilde{\bf A} + \widetilde{\bf A}^2 - \widetilde{\bf A}^4$ to depict the joint difference under the 2-order relations at the beginning. 
The power $2^{j-1}$ are utilized based on diffusion wavelets, theoretically promoting optimal localization~\cite{gama2018diffusion}.
Besides, the intuition of this design is sketched in Fig.~\ref{fig:scatterprinciple}. Suppose all the edge weights are positive, $h_0(\widetilde{\bf A})$ only preserves the low-frequency to enhance smoothness;  the other filters obtain the band-pass features and promote joints' varieties.
The real feature responses of the graph scattering during model inference is visualized in Appendix to analyze this module.

Plugging Eq.~\eqref{eq:wavelets} to Eq.~\eqref{eq:GNN_with_filter}, we output the spectrum channels $\{ {\bf H}_{(k)}\}$. 
At the next layer of the graph scattering, we repeat Eq.~\eqref{eq:GNN_with_filter} on each ${\bf H}_{(k)}$. Thus, each non-leaf feature has $K+1$ new branches; eventually, the output has $(K+1)^{L}$ channels corresponding to different spectrums.

\mypar{Adaptive graph spectrum aggregation} 
To abstract the key information from the $(K+1)^{L}$ graph scattering responses and provide information to the downstream MPGSB, we propose the adaptive graph spectrum aggregation to fuse the spectral channels based on the inferred spectrum scores, which measure the importance of each channel over the whole spectrum. Given the output channels $\{{\bf H}_{(k)}\}$, the spectrum aggregation is formulated as 
\begin{equation}
    \setlength{\abovedisplayskip}{5pt}
    \setlength{\belowdisplayskip}{5pt}
    {\bf H} = \sum\nolimits_{k=0}^{(K+1)^L} {\omega}_{k} {\bf H}_{(k)} \in \mathbb{R}^{M' \times C'},
\end{equation}
where $\omega_k$ is the inferred spectrum importance score of the $k$th feature ${\bf H}_{(k)}$, which is computed through
\begin{equation}
    \setlength{\abovedisplayskip}{5pt}
    \setlength{\belowdisplayskip}{5pt}
    {\omega}_k =
    \frac
    {\exp \left( {f}_{\rm 2} \left( {\rm tanh} \left( {f}_{\rm 1} \left( [{\bf H}_{\rm sp},{\bf H}_{(k)}] \right) \right) \right) \right)}
    {\sum_{j=0}^{(K+1)^{L}} \exp \left( {f}_{\rm 2} \left ( {\rm tanh} \left( {f}_{\rm 1} \left( [{\bf H}_{\rm sp},{\bf H}_{(j)}] \right) \right) \right) \right)},
\end{equation}
where $f_1(\cdot)$ and $f_2(\cdot)$ are MLPs, and $[\cdot,\cdot]$ is concatenation along feature dimensions.
${{\bf H}_{\rm sp}}\in\mathbb{R}^{M' \times C'}$ carries the whole graph spectrum, which is
\begin{equation}
    \setlength{\abovedisplayskip}{5pt}
    \setlength{\belowdisplayskip}{5pt}
    {\bf H}_{\rm sp} = {\rm ReLU}\left(\frac{1}{(K+1)^L}\sum\nolimits_{k=0}^{(K+1)^L}{\bf H}_{(k)}{\bf W}_{\rm sp} \right),
\end{equation}
where ${\bf W}_{\rm sp}$ denotes trainable weights and ${\rm ReLU}(\cdot)$ is the ReLU activation. ${\bf H}_{\rm sp}$ employs the embedded spectrum to benefit the representation understanding.

% Given the learned importance scores, we then use them to weight spectral channels for aggregation. The aggregated spectrum ${\bf H}\in\mathbb{R}^{M' \times C'}$ is formulated as
% \begin{equation}
%     \setlength{\abovedisplayskip}{5pt}
%     \setlength{\belowdisplayskip}{5pt}
%     {\bf H} = \sum\nolimits_{k=0}^{(K+1)^L} {\omega}_{k} {\bf H}_{(k)},
% \end{equation}
% where a larger ${\omega}_{k}$ reflects a higher influence of the corresponding ${\bf H}_{(k)}$. Furthermore, ${\bf H}$ is fed to the next MPGSB.

\begin{figure}[t]
    \centering
    \includegraphics[width=0.8\columnwidth]{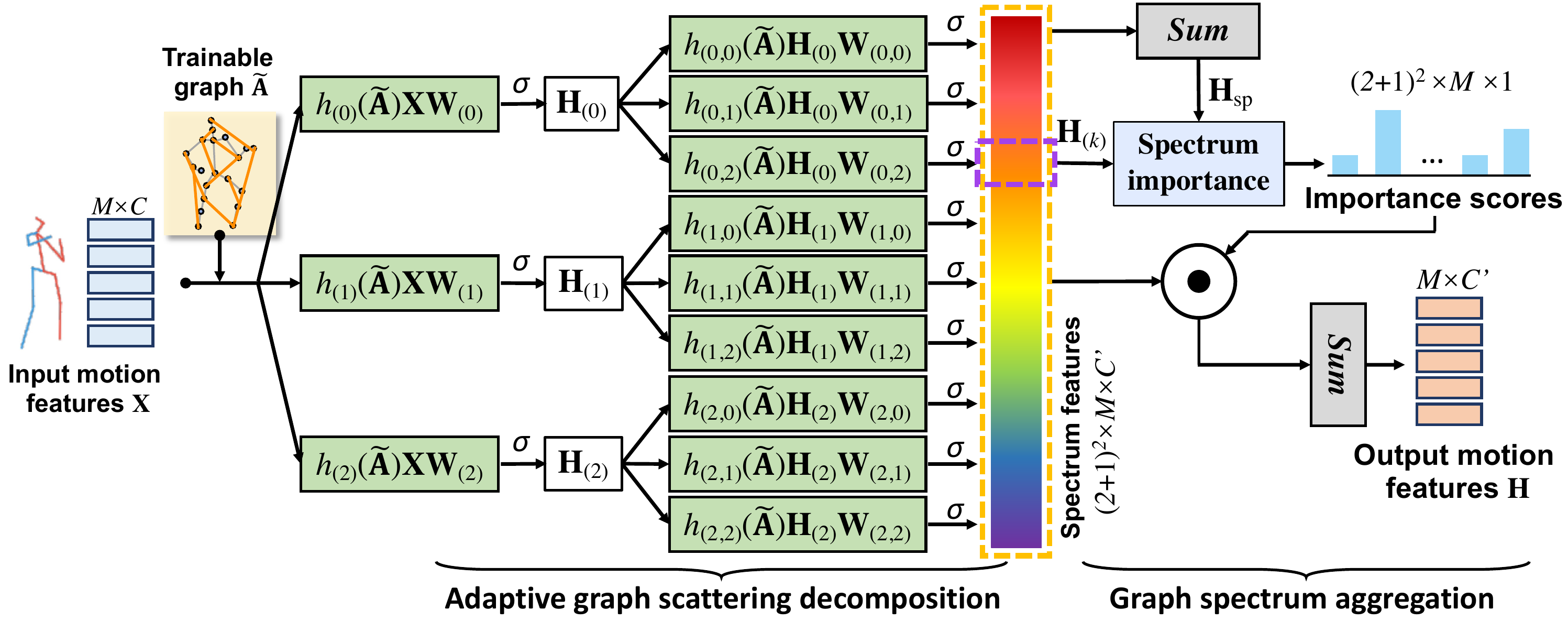}
    %\vspace{-4mm}
    \caption{\small A sketch of single-part adaptive graph scattering.}
    \label{fig:scatter}
    %\vspace{-5mm}
\end{figure}

For clearer understanding, Fig.~\ref{fig:scatter} sketches an examplar architecture of the single-part adaptive graph scattering ($L=2$ and $K=2$), where we briefly note various filtering with different subscripts.
For multiple body-parts, we leverage shared parameters but different pose graphs on them to reduce the model complexity, since different body-parts carry the same feature modality but different graph views. In this way, we can capture the diverse features that reflects rich structural information.

% Compared to existing graph scattering transforms with designed filtering~\cite{Gao_2019_ICML_GS,Ioannidis2020Pruned,NEURIPS2020_a6b964c0,pan2021spatiotemporal}, we introduce more trainable components to enhance flexibility and generalization in complex cases. Our adaptivity is four-fold. First, the adaptive graph structures exploit implicit interactions between joints; second, the adaptive filter coefficients flexibly scale graph filters; third, the adaptive network parameters capture underlying representation; last, we propose the adaptive graph spectrum aggregation, which measures the importance of individual scattering channels over the spectrum to adaptive collect the key information.

%\vspace{-2mm}
\subsection{Bipartite Cross-Part Fusion}
\label{sec:Two_Fuse}
%\vspace{-1mm}
To combine the diverse and hybrid features learned from different body-parts, we propose a bipartite cross-part fusion module based on body-part interactions, which allow the isolated single-part features to adapt to each other, leading to more coordinated and reasonable patterns on the whole body.

In this paper, we mainly consider to separate the body into two parts besides the entire body, because according to the experiences, the separation of the two parts (e.g. separating as upper and lower bodies or left and right bodies) can distinguish the different global movement patterns and explore their interactions to reflect the movement coordination. Meanwhile, the two-part separation needs only one interaction-based fusion, which reduces the model complexity. As an example, we consider the upper-lower-separation. We first model their cross-part influence, to reflect the implicit upper-lower interactions. Two directed bipartite graphs are adaptively constructed to propagate influence from the upper to the lower bodies and vice versa. Here we present the `upper-to-lower' graph as an example. Let the joint features on the upper body and lower body be ${\bf H}_{\uparrow}\in\mathbb{R}^{M_{\uparrow} \times C'}$ and ${\bf H}_{\downarrow}\in\mathbb{R}^{M_{\downarrow} \times C'}$, respectively, where $M_{\uparrow}$ and $M_{\downarrow}$ are the numbers of nodes in these two parts; we calculate the upper-to-lower affinity matrix through
\begin{equation}
    \setlength{\abovedisplayskip}{5pt}
    \setlength{\belowdisplayskip}{5pt}
    {\bf A}_{\uparrow 2 \downarrow} = {\rm softmax}(f_{\uparrow}({\bf H}_{\uparrow})f_{\downarrow}({\bf H}_{\downarrow})^{\top}) \in [0,1]^{M_{\uparrow} \times M_{\downarrow}},
\end{equation}
where ${\rm softmax}(\cdot)$ is the softmax function across rows to normalize the affinity effects and enhance the strong correlations; $f_{\uparrow}(\cdot)$ and $f_{\downarrow}(\cdot)$ are two embedding networks. Each column of ${\bf A}_{\uparrow 2 \downarrow}$ reflects the influence levels of all the upper-body-joints to the corresponding lower-body-joint.

Given the ${\bf A}_{\uparrow 2 \downarrow}$, we update the lower body via
\begin{equation}
    \setlength{\abovedisplayskip}{5pt}
    \setlength{\belowdisplayskip}{5pt}
    {\bf H}_{\downarrow}' = {\bf H}_{\downarrow} + {\bf A}_{\uparrow 2 \downarrow}^{\top}{\bf H}_{\uparrow},
\end{equation}
where the new lower body aggregates the information from the upper body by emphasizing the influence across body-parts.
In the similar manner, we also update the upper body based on the bipartite lower-to-upper graph.

Finally, given the updated upper and lower body, ${\bf H}_{\uparrow}'$ and ${\bf H}_{\downarrow}'$, we fuse them on the whole body and obtain the hybrid feature ${\bf H}'\in \mathbb{R}^{M' \times C'}$ by
\begin{equation}
    \setlength{\abovedisplayskip}{5pt}
    \setlength{\belowdisplayskip}{5pt}
    {\bf H}' = {\rm MLP} ({\bf H} + ({\bf H}_{\uparrow}' \oplus {\bf H}_{\downarrow}')),
\end{equation}
where $\oplus:\mathbb{R}^{M_{\uparrow} \times C'} \times \mathbb{R}^{M_{\downarrow} \times C'} \to \mathbb{R}^{M' \times C'}$ places joints from different body-parts to align with the original body. ${\rm MLP}(\cdot)$ further embeds the fused body. In this way, the output features carry the comprehensive graph spectrum and multi-part representation to promote motion prediction.

%\vspace{-2mm}
\subsection{Loss Function}
%\vspace{-1mm}
To train the proposed SPGSN, we define the loss function. Suppose that we take $N$ samples in a mini-batch as inputs, and let the $n$th ground-truth and predicted motion sample be $\mathbb{X}^{+}_n$ and $\widehat{\mathbb{X}}^{+}_n$. The loss function $\mathcal{L}$ is defined as the average $\ell_2$ distance between the targets and predictions:
\begin{equation}
    \setlength{\abovedisplayskip}{5pt}
    \setlength{\belowdisplayskip}{5pt}
    \mathcal{L} = \frac{1}{N}\sum\nolimits_{n=1}^{N}\|\mathbb{X}^{+}_n-\widehat{\mathbb{X}}^{+}_n\|^2.
\end{equation}
All the trainable parameters in our SPGSN are tuned end-to-end, including the body graph structures, adaptive filter coefficients and the network weights.

%\vspace{-2mm}
\section{Experiments}
\label{sec:One_exper}
%\vspace{-2mm}
\subsection{Datasets}
%\vspace{-1mm}
\label{sec:Two_dataset}
\mypar{Dataset 1: Human 3.6M (H3.6M)}
There are $7$ subjects performing $15$ classes of actions in H3.6M~\cite{ionescu2013human3}, and each subject has $22$ body joints. All sequences are downsampled by two along time. Following previous paradigms~\cite{Dang_2021_ICCV}, the models are trained on the segmented clips in the $6$ subjects and tested on the clips in the $5$th subject.

\noindent \mypar{Dataset 2: CMU Mocap}
CMU Mocap consists of $5$ general classes of actions. On each pose, we use $25$ joints in the 3D space. Following~\cite{Mao_2019_ICCV,Dang_2021_ICCV}, we use $8$ actions: `basketball', `basketball signal', `directing traffic', `jumping', `running', `soccer', `walking' and `washing window'.

\noindent \mypar{Dataset 3: 3D Pose in the Wild (3DPW)}
3DPW~\cite{Marcard_2018_ECCV} contains more than 51k frames with 3D poses for indoor and outdoor activities. We adopt the training, test and validation separation suggested by the official setting. Each subject has $23$ joints. The frame rate of the motions is $30$Hz.

%\vspace{-2mm}
\subsection{Model and Experimental Settings}
\label{sec:Two_baselines}
%\vspace{-1mm}
\mypar{Implementation details}
We implement SPGSN with PyTorch 1.4 on one NVIDIA Tesla V100 GPU. We set $10$ MPGSBs to form the entire model. In each MPGSB, the single-part adaptive graph scattering has $L=2$ layers of graph scattering decomposition, and the filter order $K=2$. The hidden dimension in a MPGSB is $256$. We use Adam optimizer~\cite{kingma2014adam} to train the SPGSN with batch size $32$. The learning rate is $0.001$ with a $0.96$ decay for every two epochs. To obtain more generalized evaluation with lower test bias, we utilize all the clips in the $5$th subject of H3.6M and the test folder of CMU Mocap, instead of testing on a few samples picked from the test sequences like in~\cite{Martinez_2017_CVPR,Li_2018_CVPR,Gui_2018_ECCV,Mao_2019_ICCV,Li_2020_CVPR}.

\noindent \textbf{Baselines.}
We compare our model to many state-of-the-art methods, including the RNN-based Res-sup~\cite{Martinez_2017_CVPR}, feed-fordward-based CSM~\cite{Li_2018_CVPR}, SkelNet~\cite{guo2019human}, and graph-based Traj-GCN~\cite{Mao_2019_ICCV}, DMGNN~\cite{Li_2020_CVPR}, HisRep~\cite{Mao_2020_ECCV}, STSGCN~\cite{Sofianos_2021_ICCV} and MSR-GCN~\cite{Dang_2021_ICCV}. We test these methods under the same protocol.

\noindent \textbf{Evaluation Metrics.}
We use the Mean Per Joint Position Error (MPJPE), where we record the average $\ell_2$ distance between predicted joints and target ones in 3D Euclidean space at each prediction timestamp. 
Compared to previous mean angle error (MAE)~\cite{Martinez_2017_CVPR,Li_2020_CVPR}, the MPJPE relfects larger degrees of freedom of human poses and covers larger ranges of errors for clearer comparison. 

\begin{table*}[t]
    \centering
    \small
    \caption{\small Prediction MPJPEs of various models for short-term motion prediction on 5 representative actions in H3.6M. We also introduce an SPGSN variant called SPGSN (1body), which only considers the entire non-separated bodies. Since the original STSGCN~\cite{Sofianos_2021_ICCV} uses a different protocol from all the other methods, we update its code for a fair comparison; see results in STSGCN*.}
    %\vspace{-2mm}
    \renewcommand{\arraystretch}{1.0}
    \resizebox{1\textwidth}{!}{
    \begin{tabular}{|c|cccc|cccc|cccc|cccc|cccc|}
        \hline
        Motion & \multicolumn{4}{c|}{Walking} & \multicolumn{4}{c|}{Eating} & \multicolumn{4}{c|}{Smoking} & \multicolumn{4}{c|}{Discussion} & \multicolumn{4}{c|}{Directions} \\
        % \hline
        millisecond & 80&160&320&400 & 80&160&320&400 & 80&160&320&400 & 80&160&320&400 & 80&160&320&400 \\
        \hline
        Res-sup~\cite{Martinez_2017_CVPR} & 29.36 & 50.82 & 76.03 & 81.52 & 16.84 & 30.60 & 56.92 & 68.65 & 22.96 & 42.64 & 70.24 & 83.68 & 32.94 & 61.18 & 90.92 & 96.19 & 35.36 & 57.27 & 76.30 & 87.67 \\
        CSM~\cite{Li_2018_CVPR} & 21.70 & 43.56 & 66.29 & 75.48 & 14.50 & 26.13 & 47.47 & 55.63 & 19.42 & 37.70 & 62.49 & 68.55 & 26.35 & 53.41 & 79.12 & 83.01 & 27.07 & 44.72 & 63.94 & 75.37 \\
        SkelNet~\cite{guo2019human} & 20.49 & 34.36 & 59.64 & 68.76 & 11.80 & 22.38 & 39.88 & 48.11 & 11.33 & 23.71 & 45.30 & 52.85 & 21.79 & 40.24 & 65.93 & 77.91 & 16.06 & 27.12 & 62.97 & 72.75 \\
        DMGNN~\cite{Li_2020_CVPR} & 17.32 & 30.67 & 54.56 & 65.20 & 10.96 & 21.39 & 36.18 & 43.88 & 8.97 & 17.62 & 32.05 & 40.30 & 17.33 & 34.78 & 61.03 & 69.80 & 13.14 & 24.62 & 64.68 & 81.86 \\
        Traj-GCN~\cite{Mao_2019_ICCV} & 12.29 & 23.03 & 39.77 & 46.12 & 8.36 & 16.90 & 33.19 & 40.70 & 7.94 & 16.24 & 31.90 & 38.90 & 12.50 & 27.40 & 58.51 & 71.68 & 8.97 & 19.87 & 43.35 & 53.74 \\
        HisRep~\cite{Mao_2020_ECCV} & 10.53 & 19.96 & 34.88 & {42.05} & 7.39 & 15.53 & 31.26 & 38.58 & 7.17 & 14.54 & 28.83 & 35.67 & 10.89 & 25.19 & 56.15 & 69.30 & 7.77 & 18.23 & 41.34 & 51.61 \\
        MSR-GCN~\cite{Dang_2021_ICCV} & 12.16 & 22.65 & 38.64 & 45.24 & 8.39 & 17.05 & 33.03 & 40.43 & 8.02 & 16.27 & 31.32 & 38.15 & 11.98 & 26.76 & 57.08 & 69.74 & 8.61 & 19.65 & 43.28 & 53.82 \\
        STSGCN*~\cite{Sofianos_2021_ICCV} & 16.26 & 24.63 & 40.06 & 45.94 & 14.32 & 22.14 & 37.91 & 45.03 & 13.10 & 20.20 & 37.71 & 44.65 & 14.33 & 24.28 & {\bf 52.62} & {68.53} & 14.24 & 24.27 & 44.24 & 53.21 \\
        \hline
        SPGSN (1body) & {\bf 10.13} & 19.51 & 35.52 & 44.67 & 7.13 & 15.02 & 31.87 & 41.18 & 6.83 & 13.94 & 28.77 & 36.78 & 10.42 & 23.90 & 54.13 & 69.99 & 7.38 & 17.48 & 40.54 & 53.09 \\
        SPGSN & 10.14 & {\bf 19.39} & {\bf 34.80} & {\bf 41.47} & {\bf 7.07} & {\bf 14.85} & {\bf 30.48} & {\bf 37.91} & {\bf 6.72} & {\bf 13.79} & {\bf 27.97} & {\bf 34.61} & {\bf 10.37} & {\bf 23.79} & 53.61 & {\bf 67.12} & {\bf 7.35} & {\bf 17.15} & {\bf 39.80} & {\bf 50.25} \\
        \hline
    \end{tabular}}
    %\vspace{-2mm}
    \label{tab:h36m_short4_3D}
\end{table*}

\begin{table*}[t]
    \centering
    \small
    \caption{\small MPJPEs for short-term motion prediction on other $9$ actions in H3.6M.}
    %\vspace{-2mm}
    \renewcommand{\arraystretch}{1.0}
    \resizebox{1\textwidth}{!}{
    \begin{tabular}{|c|cccc|cccc|cccc|cccc|cccc|}
        \hline
        Motion & \multicolumn{4}{c|}{Greeting} & \multicolumn{4}{c|}{Phoning} & \multicolumn{4}{c|}{Posing} & 
        \multicolumn{4}{c|}{Purchases} & \multicolumn{4}{c|}{Sitting} \\
        % \hline
        millisecond & 80&160&320&400 & 80&160&320&400 & 80&160&320&400 & 80&160&320&400 & 80&160&320&400 \\
        \hline
        Res-sup~\cite{Martinez_2017_CVPR} & 34.46 & 63.36 & 124.60 & 142.50 & 37.96 & 69.32 & 115.00 & 126.73 & 36.10 & 69.12 & 130.46 & 157.08 & 36.33 & 60.30 & 86.53 & 95.92 & 42.55 & 81.40 & 134.70 & 151.78  \\
        DMGNN~\cite{Li_2020_CVPR} & 23.30 & 50.32 & 107.30 & 132.10 & 12.47 & 25.77 & 48.08 & 58.29 & 15.27 & 29.27 & 71.54 & 96.65 & 21.35 & 38.71 & 75.67 & 92.74 & 11.92 & 25.11 & 44.59 & {\bf 50.20} \\
        Traj-GCN~\cite{Mao_2019_ICCV} & 18.65 & 38.68 & 77.74 & 93.39 & 10.24 & 21.02 & 42.54 & 52.30 & 13.66 & 29.89 & 66.62 & 84.05 & 15.60 & 32.78 & 65.72 & 79.25 & 10.62 & 21.90 & 46.33 & 57.91  \\
        MSR-GCN~\cite{Dang_2021_ICCV} & 16.48 & 36.95 & 77.32 & 93.38 & 10.10 & 20.74 & 41.51 & 51.26 & 12.79 & 29.38 & 66.95 & 85.01 & 14.75 & 32.39 & 66.13 & 79.64 & 10.53 & 21.99 & 46.26 & 57.80 \\
        STSGCN*~\cite{Sofianos_2021_ICCV} & 15.02 & {\bf 30.70} & {\bf 67.11} & {87.63} & 14.88 & 21.40 & 46.55 & 52.03 & 15.01 & 25.69 & {\bf 58.38} & {\bf 73.08} & 15.26 & {\bf 26.26} & {63.45} & {\bf 74.25} & 15.19 & 22.95 & 46.82 & 58.34  \\
        \hline
        SPGSN (1body) & 15.16 & 33.61 & 71.89 & 88.74 & 8.78 & 18.50 & 39.85 & 51.53 & 10.92 & 25.46 & 61.38 & 78.87 & 12.78 & 28.86 & 62.59 & 77.01 & {\bf 9.25} & 19.58 & 43.47 & 56.32  \\
        SPGSN & {\bf 14.64} & 32.59 & 70.64 & {\bf 86.44} & {\bf 8.67} & {\bf 18.32} & {\bf 38.73} & {\bf 48.46} & {\bf 10.73} & {\bf 25.31} & 59.91 & 76.46 & {\bf 12.75} & 28.58 & {\bf 61.01} & 74.38 & 9.28 & {\bf 19.40} & {\bf 42.25} & {53.56}  \\
        \hline
        Motion & \multicolumn{4}{c|}{Sitting Down} & \multicolumn{4}{c|}{Taking Photo} & \multicolumn{4}{c|}{Waiting} & \multicolumn{4}{c|}{Walking Together} & \multicolumn{4}{c|}{Average} \\
        % \hline
        millisecond  & 80&160&320&400 & 80&160&320&400 & 80&160&320&400 & 80&160&320&400 & 80&160&320&400 \\
        \cline{1-21}
        Res-sup~\cite{Martinez_2017_CVPR} & 47.28 & 85.95 & 145.75 & 168.86 & 26.10 & 47.61 & 81.40 & 94.73 & 30.62 & 57.82 & 106.22 & 121.45 & 26.79 & 50.07 & 80.16 & 92.23 & 34.66 & 61.97 & 101.08 & 115.49 \\
        DMGNN~\cite{Li_2020_CVPR} & 14.95 & 32.88 & 77.06 & 93.00 & 13.61 & 28.95 & 45.99 & 58.76 & 12.20 & 24.17 & 59.62 & 77.54 & 14.34 & 26.67 & 50.08 & 63.22 & 16.95 & 33.62 & 65.90 & 79.65 \\
        Traj-GCN~\cite{Mao_2019_ICCV} & 16.14 & 31.12 & 61.47 & 75.46 & 9.88 & 20.89 & 44.95 & 56.58 & 11.43 & 23.99 & 50.06 & 61.48 & 10.47 & 21.04 & 38.47 & 45.19 & 12.68 & 26.06 & 52.27 & 63.51 \\
        MSR-GCN~\cite{Dang_2021_ICCV} & 16.10 & 31.63 & 62.45 & 76.84 & 9.89 & 21.01 & 44.56 & 56.30 & 10.68 & 23.06 & 48.25 & 59.23 & 10.56 & 20.92 & 37.40 & 43.85 & 12.11 & 25.56 & 51.64 & 62.93 \\
        STSGCN*~\cite{Sofianos_2021_ICCV} & 16.70 & 28.05 & {\bf 56.15} & {72.03} & 16.61 & 24.84 & 45.98 & 61.79 & 16.30 & 24.33 & 48.12 & 59.79 & 11.38 & 22.39 & 39.90 & 47.48 & 15.34 & 25.52 & 50.64 & 60.61 \\
        \cline{1-21}
        SPGSN (1body) & 14.34 & 28.10 & 58.23 & 74.44 & {\bf 8.72} & 18.95 & 42.62 & 55.22 & 9.24 & 20.02 & 43.80 & 56.80 & {\bf 8.91} & 18.46 & 34.88 & 42.98 & 10.55 & 22.63 & 48.21 & 60.96 \\
        SPGSN & {\bf 14.18} & {\bf 27.72} & 56.75 & {\bf 70.74} & 8.79 & {\bf 18.90} & {\bf 41.49} & {\bf 52.66} & {\bf 9.21} & {\bf 19.79} & {\bf 43.10} & {\bf 54.14} & 8.94 & {\bf 18.19} & {\bf 33.84} & {\bf 40.88} & {\bf 10.44} & {\bf 22.33} & {\bf 47.07} & {\bf 58.26} \\
        \cline{1-21}
    \end{tabular}}
    %\vspace{-2mm}
    \label{tab:h36m_short_3D}
\end{table*}

\begin{table*}[!t]
    \centering
    \caption{\small Prediction MPJPEs of methods for long-term prediction on $8$ actions in H3.6M and the average MPJPEs across all the actions.}
    %\vspace{-2mm}
    \renewcommand{\arraystretch}{1.0}
    \footnotesize
    \resizebox{1\textwidth}{!}{
    \begin{tabular}{|c|cc|cc|cc|cc|cc|cc|cc|cc|cc|}
        \hline
        Motion & 
        \multicolumn{2}{c|}{Walking}& \multicolumn{2}{c|}{Eating}&
        \multicolumn{2}{c|}{Smoking}& 
        \multicolumn{2}{c|}{Directions} & \multicolumn{2}{c|}{Phoning} &
        \multicolumn{2}{c|}{Sitting} & \multicolumn{2}{c|}{TakingPhoto} & \multicolumn{2}{c|}{Waiting} & \multicolumn{2}{c|}{Average} \\
        millisecond & 560 & 1k & 560 & 1k & 560 & 1k & 560 & 1k & 560 & 1k & 560 & 1k & 560 & 1k & 560 & 1k & 560 & 1k\\
        \hline
        Res-sup.~\cite{Martinez_2017_CVPR} & 81.73 & 100.68 & 79.87 & 100.20 & 94.83 & 137.44 & 110.05 & 152.48 & 143.92 & 186.79 & 166.20 & 185.16 & 107.03 & 162.38 & 126.70 & 153.14 & 129.19 & 164.96 \\
        Traj-GCN~\cite{Mao_2019_ICCV} & 54.05 & 59.75 & 53.39 & 77.75 & 50.74 & 72.62 & 71.01 & 101.79 & 69.55 & 104.19 & 77.63 & 118.36 & 78.73 & 120.06 & 79.08 & 107.32 & 81.07 & 113.01 \\ 
        DMGNN~\cite{Li_2020_CVPR} & 71.36 & 95.82 & 58.11 & 86.66 & 50.85 & 72.15 & 102.06 & 135.75 & 71.33 & 108.37 & 75.51 & {\bf 115.44} & 78.38 & 123.65 & 85.54 & 113.68 & 93.57 & 127.62 \\ 
        MSR-GCN~\cite{Dang_2021_ICCV} & 52.72 & 63.05 & 52.54 & 77.11 & 49.45 & 71.64 & 71.18 & 100.59 & 68.28 & 104.36 & 78.19 & 120.02 & 77.94 & 121.87 & 76.33 & 106.25 & 81.13 & 114.18 \\
        STSGCN*~\cite{Sofianos_2021_ICCV} & 50.64 & 64.74 & 56.46 & 75.08 & 55.55 & 74.13 & 75.61 & 109.89 & 79.19 & 109.88 & 82.32 & 119.83 & 87.70 & 119.79 & 78.41 & 108.04 & 80.66 & 113.33 \\
        \hline
        SPGSN & {\bf 46.89} & {\bf 53.59} & {\bf 49.76} & {\bf 73.39} & {\bf 46.68} & {\bf 68.62} & {\bf 70.05} & {\bf 100.52} & {\bf 66.70} & {\bf 102.52} & {\bf 75.00} & 116.24 & {\bf 75.58} & {\bf 118.22} & {\bf 73.50} & {\bf 103.62} & {\bf 77.40} & {\bf 109.64} \\ 
        \hline
    \end{tabular}}
    %\vspace{-5mm}
    \label{tab:h36m_long_MPJPE}
\end{table*}

\begin{table*}[t]
    \centering
    \caption{\small Prediction MPJPEs of methods on CMU Mocap for both short-term and long-term prediction, as well as the average prediction results across all the actions.}
    %\vspace{-2mm}
    \footnotesize
    \renewcommand{\arraystretch}{1.0}
    \resizebox{1\textwidth}{!}{
        \begin{tabular}{|c|ccccc|ccccc|ccccc|ccccc|}
        \hline
        Motion & \multicolumn{5}{c|}{Basketball} & \multicolumn{5}{c|}{Basketball Signal} & \multicolumn{5}{c|}{Directing Traffic} & \multicolumn{5}{c|}{Jumping} \\ 
        % \hline
        millisecond & 80 & 160 & 320 & 400 & 1000 & 80 & 160 & 320 & 400 & 1000 & 80 & 160 & 320 & 400 & 1000 & 80 & 160 & 320 & 400 & 1000 \\ \hline
        Res-sup.~\cite{Martinez_2017_CVPR} & 15.45 & 26.88 & 43.51 & 49.23 & 88.73 & 20.17 & 32.98 & 42.75 & 44.65 & 60.57 & 20.52 & 40.58 & 75.38 & 90.36 & 153.12 & 26.85 & 48.07 & 93.50 & 108.90 & 162.84 \\
        DMGNN~\cite{Li_2020_CVPR} & 15.57 & 28.72 & 59.01 & 73.05 & 138.62 & 5.03 & 9.28 & 20.21 & 26.23 & 52.04 & 10.21 & 20.90 & 41.55 & 52.28 & 111.23 & 31.97 & 54.32 & 96.66 & 119.92 & 224.63 \\
        Traj-GCN~\cite{Mao_2019_ICCV} & 11.68 & 21.26 & 40.99 & 50.78 & 97.99 & 3.33 & 6.25 & 13.58 & 17.98 & 54.00 & 6.92 & 13.69 & 30.30 & 39.97 & 114.16 & 17.18 & 32.37 & 60.12 & 72.55 & 127.41 \\
        MST-GCN~\cite{Dang_2021_ICCV} & 10.28 & 18.94 & {\bf 37.68} & {\bf 47.03} & {\bf 86.96} & 3.03 & 5.68 & 12.35 & 16.26 & 47.91 & 5.92 & 12.09 & 28.36 & 38.04 & 111.04 & 14.99 & 28.66 & {\bf 55.86} & {\bf 69.05} & {\bf 124.79} \\
        STSGCN~\cite{Sofianos_2021_ICCV} & 12.56 & 23.04 & 41.92 & 50.33 & 94.17 & 4.72 & 6.69 & 14.53 & 17.88 & 49.52 & 6.41 & 12.38 & 29.05 & 38.86 & 109.42 & 17.52 & 31.48 & 58.74 & 72.06 & 127.40 \\
        \hline
        SPGSN & {\bf 10.24} & {\bf 18.54} & 38.22 & 48.68 & 89.58 & {\bf 2.91} & {\bf 5.25} & {\bf 11.31} & {\bf 15.01} & {\bf 47.31} & {\bf 5.52} & {\bf 11.16} & {\bf 25.48} & {\bf 37.06} & {\bf 108.14} & {\bf 14.93} & {\bf 28.16} & 56.72 & 71.16 & 125.20 \\
        \hline
        %\cline{1-21}
        Motion & \multicolumn{5}{c|}{Soccer} & \multicolumn{5}{c|}{Walking} & \multicolumn{5}{c|}{Washing Window} & \multicolumn{5}{c|}{Average} \\ 
        % \cline{1-21}
        millisecond & 80 & 160 & 320 & 400 & 1000 & 80 & 160 & 320 & 400 & 1000 & 80 & 160 & 320 & 400 & 1000 & 80 & 160 & 320 & 400 & 1000 \\ 
        \cline{1-21}
        Res-sup.~\cite{Martinez_2017_CVPR} & 17.75 & 31.30 & 52.55 & 61.40 & 107.37 & 44.35 & 76.66 & 126.83 & 151.43 & 194.33 & 22.84 & 44.71 & 86.78 & 104.68 & 202.73 & 24.21 & 43.75 & 76.19 & 88.93 & 139.00 \\
        DMGNN~\cite{Li_2020_CVPR} & 14.86 & 25.29 & 52.21 & 65.42 & 111.90 & 9.57 & 15.53 & 26.03 & 30.37 & 67.01 & 7.93 & 14.68 & 33.34 & 44.24 & 82.84 & 14.07 & 24.44 & 45.90 & 55.45 & 104.33 \\
        Traj-GCN~\cite{Mao_2019_ICCV} & 13.33 & 24.00 & 43.77 & 53.20 & 108.26 & 6.62 & 10.74 & 17.40 & 20.35 & {\bf 34.41} & 5.96 & 11.62 & 24.77 & 31.63 & 66.85 & 9.94 & 18.02 & 33.55 & 40.95 & 81.85 \\
        MSR-GCN~\cite{Dang_2021_ICCV} & 10.92 & 19.50 & 37.05 & 46.38 & {\bf 99.32} & {\bf 6.31} & 10.30 & 17.64 & 21.12 & 39.70 & 5.49 & 11.07 & 25.05 & 32.51 & 71.30 & 8.72 & 15.83 & 30.57 & 38.10 & 79.01 \\
        STSGCN~\cite{Sofianos_2021_ICCV} & 13.49 & 25.24 & 39.87 & 51.58 & 109.63 & 7.18 & 10.99 & 17.84 & 22.61 & 44.12 & 6.79 & 12.10 & 24.92 & 36.66 & 69.48 & 10.80 & 18.19 & 31.18 & 41.05 & 81.76\\
        \cline{1-21}
        SPGSN & {\bf 10.86} & {\bf 18.99} & {\bf 35.05} & {\bf 45.16} & 99.51 & 6.32 & {\bf 10.21} & {\bf 16.34} & {\bf 20.19} & 34.83 & {\bf 4.86} & {\bf 9.44} & {\bf 21.50} & {\bf 28.37} & {\bf 65.08} & {\bf 8.30} & {\bf 14.80} & {\bf 28.64} & {\bf 36.96} & {\bf 77.82} \\ 
        \cline{1-21}
    \end{tabular}}
    %\vspace{-3mm}
    \label{tab:pred_cmu}
\end{table*}

\begin{table}[t]
      \small 
      \centering
      \caption{\small The average prediction MPJPEs across the test set of 3DPW at various prediction time steps.}
      %\vspace{-2mm}
      \renewcommand{\arraystretch}{1.0}
      \resizebox{0.7\columnwidth}{!}{
      \begin{tabular}{|c|cccccccc|}
          \hline
          ~ & \multicolumn{8}{c|}{Average MAE} \\
          \hline
          millisecond & 100 & 200 & 400 & 500 & 600 & 800 & 900 & 1000 \\
          \hline
          Res-sup.~\cite{Martinez_2017_CVPR} & 102.28 & 113.24 & 173.94 & 185.35 & 191.47 & 201.39 & 205.12 & 210.58 \\
          CSM~\cite{Li_2018_CVPR} & 57.83 & 71.53 & 124.01 & 142.47 & 155.16 & 174.87 & 183.40 & 187.06 \\
          Traj-GCN~\cite{Mao_2019_ICCV} & 16.28 & 35.62 & 67.46 & 80.19 & {\bf 90.36} & 106.79 & 113.93 & 117.84 \\
          DMGNN~\cite{Li_2020_CVPR} & 17.80 & 37.11 & 70.38 & 83.02 & 94.12 & 109.67 & 117.25 & 123.93  \\
          HisRep~\cite{Mao_2020_ECCV} & 15.88 & 35.14 & 66.82 & 78.49 & 93.55 & 107.63 & 114.59 & 114.75 \\
          MSR-GCN~\cite{Dang_2021_ICCV} & 15.70 & 33.48 & 65.02 & 77.59 & 93.81 & 108.15 & 114.88 & 116.31 \\
          STSGCN~\cite{Sofianos_2021_ICCV} & 18.32 & 37.79 & 67.51 & 77.34 & 92.75 & 106.65 & 113.14 & 112.22 \\
          \hline
          SPGSN & {\bf 15.39} & {\bf 32.91} & {\bf 64.54} & {\bf 76.23} & 91.62 & {\bf 103.98} & {\bf 109.41} & {\bf 111.05} \\
          \hline
      \end{tabular}}
      %\vspace{-5mm}
      \label{tab:pred_3DPW}
\end{table}

\begin{figure}[t]
    \centering
    \includegraphics[width=0.74\columnwidth]{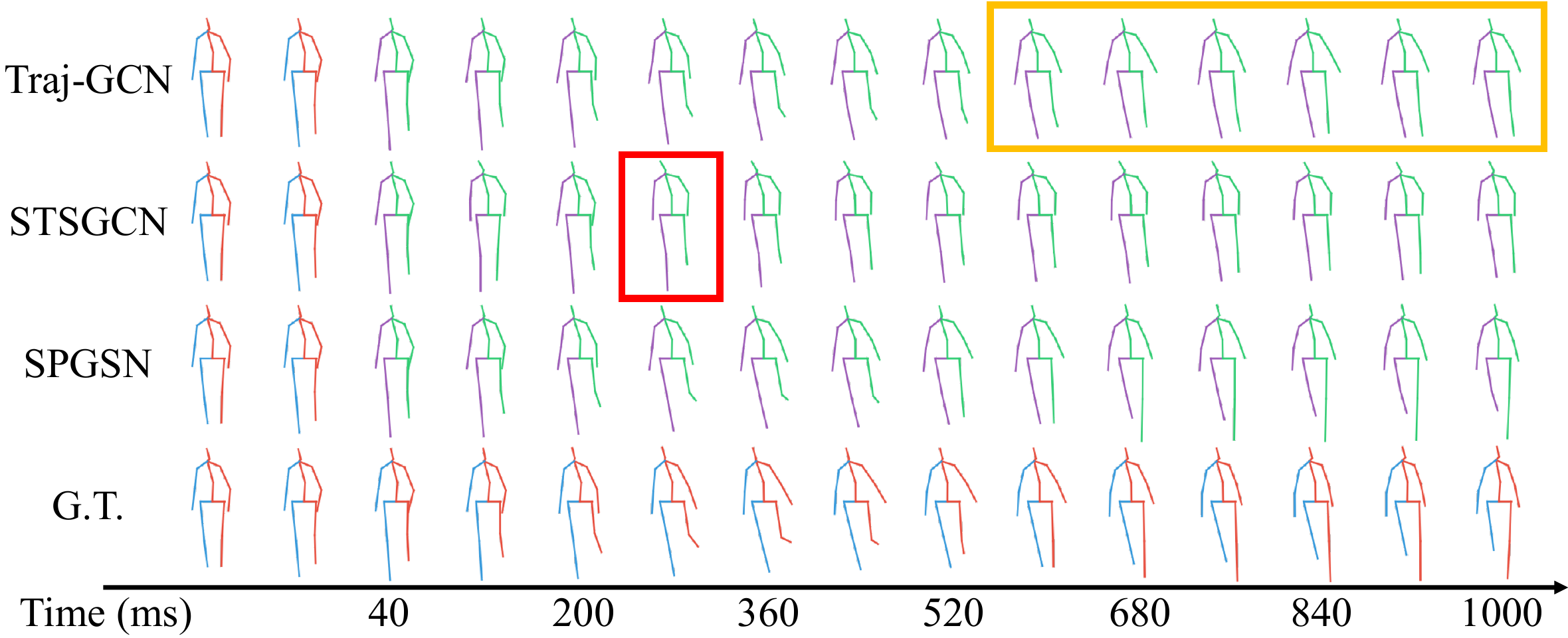}
    %\vspace{-3mm}
    \caption{\small Prediction samples of different methods on action `Walking' of H3.6M for long-term prediction. The predictions of Traj-GCN collapse to static `mean poses' after the 600th ms (orange box); STSGCN starts to suffer from large errors at the 280th ms (red box); SPGSN completes the action more accurately.}
    \label{fig:PredictionSample}
    %\vspace{-1mm}
\end{figure}

\begin{table}[t]
      \small 
      \centering
      \caption{\small Performance analysis of SPGSNs with different numbers of graph scattering blocks and graph scattering layers.}
      % \vspace{-2mm}
      \renewcommand{\arraystretch}{1.0}
      \resizebox{0.65\columnwidth}{!}{
      \begin{tabular}{|c|ccc|cccc|cc|}
          \hline
          ~ & \multicolumn{3}{c|}{Scatter-layers} & \multicolumn{6}{c|}{Average MPJPE} \\
          \hline
          MPGSB number & 1 & 2 & 3 & 80 & 160 & 320 & 400 & 560 & 1000 \\
          \hline 
          \multirow{3}{*}{8}  & \Checkmark & & & 11.34 & 23.86 & 49.66 & 61.22 & 80.32 & 114.26 \\
          ~                   & & \Checkmark & & 11.05 & 23.27 & 48.58 & 59.97 & 78.81 & 112.29 \\
          ~                   & & & \Checkmark & 10.88 & 23.12 & 48.43 & 59.76 & 79.32 & 112.86 \\
          \hline
          \multirow{3}{*}{9}  & \Checkmark & & & 10.91 & 23.22 & 48.39 & 59.63 & 79.31 & 113.09 \\
          ~                   & & \Checkmark & & 10.58 & 22.56 & 47.49 & 58.73 & 78.49 & 112.06 \\
          ~                   & & & \Checkmark & {\bf 10.40} & 22.33 & 47.29 & 58.58 & 77.79 & 110.55 \\
          \hline
          \multirow{3}{*}{10} & \Checkmark & & & 10.48 & 22.41 & 47.53 & 58.78 & 78.81 & 111.29 \\
          ~                   & & \Checkmark & & {10.44} & {\bf 22.33} & {\bf 47.07} & {\bf 58.26} & {\bf 77.40} & {\bf 109.64} \\
          ~                   & & & \Checkmark & 10.46 & 22.46 & 47.25 & 58.48 &  77.86 & 110.64\\
          \hline
          \multirow{3}{*}{11} & \Checkmark & & & 10.51 & 22.39 & 47.24 & 58.42 & 77.91 & 110.35 \\
          ~                   & & \Checkmark & & 10.57 & 22.53 & 47.15 & 58.49 & 79.06 &	112.05 \\
          ~                   & & & \Checkmark & 10.66 & 22.69 & 47.43 & 58.66 & 79.18 & 112.83 \\
          \hline
      \end{tabular}}
      %\vspace{-3mm}
      \label{tab:ablation_blocks}
\end{table}

\begin{table}[!t]
      \small 
      \centering
      \caption{\small SPGSNs with different numbers of graph scattering channels on each non-leaf scattering feature.}
      % \vspace{-2mm}
      \renewcommand{\arraystretch}{1.0}
      \resizebox{0.65\columnwidth}{!}{
      \begin{tabular}{|c|cccc|cc|}
          \hline
          ~ & \multicolumn{6}{c|}{Average MAE} \\
          \hline
          channel numbers ($K$+$1$) & 80 & 160 & 320 & 400 & 560 & 1000 \\
          \hline
          1 & 12.03 & 25.87 & 51.40 & 61.98 & 80.52 & 113.15 \\
          2 & 10.50 & 22.39 & 47.40 & 58.44 & 77.53 & 110.35 \\
          3 & {\bf 10.44} & {\bf 22.33} & {\bf 47.07} & {\bf 58.26} & 77.40 & {\bf 109.64} \\
          4 & 10.47 & 22.41 & 47.23 & 58.32 & {\bf 77.37} & 109.89 \\
          5 & 10.57 & 22.46 & 47.38 & 58.46 & 77.66 & 110.52 \\
          \hline
      \end{tabular}}
      % \vspace{-5mm}
      \label{tab:ablation_channles}
\end{table}

\begin{table}[t]
    \small
    \centering
    \caption{Comparison between SPGSN and its variants, including the model without cross-part fusion, i.e., SPGSN (no CrossPart), and the model separating the body into left and right parts, i.e., SPGSN (left-right).}
    % \vspace{-2mm}
    \resizebox{0.7\columnwidth}{!}{
    \begin{tabular}{|c|cccc|cc|}
        \hline
        prediction time & 80 & 160 & 320 & 400 & 560 & 1000 \\ 
        \hline
        SPGSN (no CrossPart) & 10.55 & 22.63 & 48.21 & 60.96 & 78.02 & 111.74 \\
        SPGSN (left-right) & 10.47 & 22.51 & 47.48 & 58.85 & 77.79 & 110.16 \\
        \hline
        SPGSN (ours, upper-lower) & {\bf 10.44} & {\bf 22.33} & {\bf 47.07} & {\bf 58.26} & {\bf 77.40} & {\bf 109.64} \\
        \hline
    \end{tabular}}
    \label{tab:scatter_or_not}
    % \vspace{-3mm}
\end{table}

%\vspace{-2mm}
\subsection{Comparison to State-of-the-Art Methods}
%\vspace{-1mm}
To validate SPGSN, we show the quantitative results for both short-term and long-term motion prediction on H3.6M, CMU Mocap and 3DPW. We also illustrate the predicted samples for qualitative evaluation.

\textbf{Short-term prediction.}
Short-term motion prediction aims to predict the poses within 400 milliseconds. First, on H3.6M, Table~\ref{tab:h36m_short4_3D} presents the MPJPEs of SPGSN and many previous methods on $5$ representative actions at multiple prediction timestamps. We see that, SPGSN obtains superior performance at most timestamps; also, learning the diverse patterns from separated body-parts, SPGSN outperforms the variant SPGSN (1body) that only uses the entire human body.
Besides, Table~\ref{tab:h36m_short_3D} presents the MPJPEs on other $9$ actions in H3.6M and the average MPJPEs over the dataset. Compared to the baselines, SPGSN achieves much lower MPJPEs by $9.3\%$ in average.
Notably, the original STSGCN~\cite{Sofianos_2021_ICCV} uses a different protocol from all the other methods, we update its code for a fair comparison; see STSGCN* and more details are in Appendix.

\textbf{Long-term prediction.}
Long-term motion prediction aims to predict the poses over 400 milliseconds, which is challenging due to the pose variation and elusive human intention. {Table~\ref{tab:h36m_long_MPJPE} presents the prediction MPJPEs of various methods at the 560 ms and 1000 ms on $8$ actions in H3.6M.} We see that, SPGSN achieves more effective prediction on most actions and has lower MPJPEs by $3.6\%$ in average. The results on the other actions are shown in Appendix.

We also test the SPGSN for both short-term and long-term prediction on CMU Mocap.  Table~\ref{tab:pred_cmu} shows the MPJPEs on $7$ actions within the future 1000 ms. We see that, SPGSN outperforms the baselines on most actions, and the average prediction MPJPE is much lower by $9.3\%$ than previous methods.

Furthermore, we test the SPGSN on 3DPW dataset for both short-term and long-term motion prediction. We present the average MPJPEs across all the test samples at different prediction steps in Table~\ref{tab:pred_3DPW}.
Compared to the state-of-the-art methods, SPGSN reduces the MPJPE by $2.7\%$ in average.

\mypar{Prediction Visualization}
To qualitatively evaluate the prediction, we compare the synthesized samples of SPGSN to those of Traj-GCN and STSGCN on H3.6M. Figure.~\ref{fig:PredictionSample} illustrates the future poses of ‘Walking’ in 1000 ms with the frame interval of 80 ms. Compared to the baselines, SPGSN completes the action more accurately. The predictions of Traj-GCN collapse to static `mean poses' after the 600th ms (orange box). STSGCN starts to suffer from large errors at the 280th ms (red box). See more results in Appendix.

%\vspace{-2mm}
\subsection{Model Analysis}
%\vspace{-1mm}
% Here we study the  effects of various configurations of the proposed SPGSN.

\mypar{Numbers of MPGSBs and graph scattering layers} We test the SPGSN frameworks with various numbers of MPGSBs ($8$-$11$) and layers of adaptive graph scattering ($1$-$3$) on H3.6M for both short-term and long-term prediction. Table~\ref{tab:ablation_blocks} presents the average MPJPEs of different architectures. The SPGSN with $10$ MPGSBs and $2$ layers of adaptive graph scattering obtains the lowest MPJPEs. The models have stable prediction for $9$-$11$ MSGCBs, and $2$ or $3$ graph scattering layers usually show better results than only using one layer.

\textbf{Numbers of spectral channels.}
We investigate the numbers of graph scattering decomposition channels, $K$, in each MPGSB. We test the model that applies $1$ to $5$ channels on H3.6M; see the average MPJPEs in Table~\ref{tab:ablation_channles}. We see that $3$ channels lead to the lowest prediction errors. The model with only $1$ channel cannot capture the sufficiently large spectrums and rich information. Five channels cause much heavy model and over-fitting.

\mypar{Bipartite cross-part interaction}
To investigate the proposed body-part separation and cross-part interaction, we compare the SPGSN with two model variants. First, SPGSN that directly aligns the independent part features on the entire body; second, SPGSN separates the body into left and right parts. We test these variants on H3.6M; see Table~\ref{tab:scatter_or_not}.
The SPGSN with upper-lower interaction and fusion consistently outperforms the baselines. For SPGSN (no CrossPart), due to the lack of mutual influence, it is hard to ensure higher coordination and rationality. As for SPGSN (left-right),  given the interaction of body-parts, the prediction error is reduced, while the upper-lower separation promotes more special dynamics than the left-right separation due to the movement diversity.

\begin{figure}[t]
    \centering
    \includegraphics[width=0.9\columnwidth]{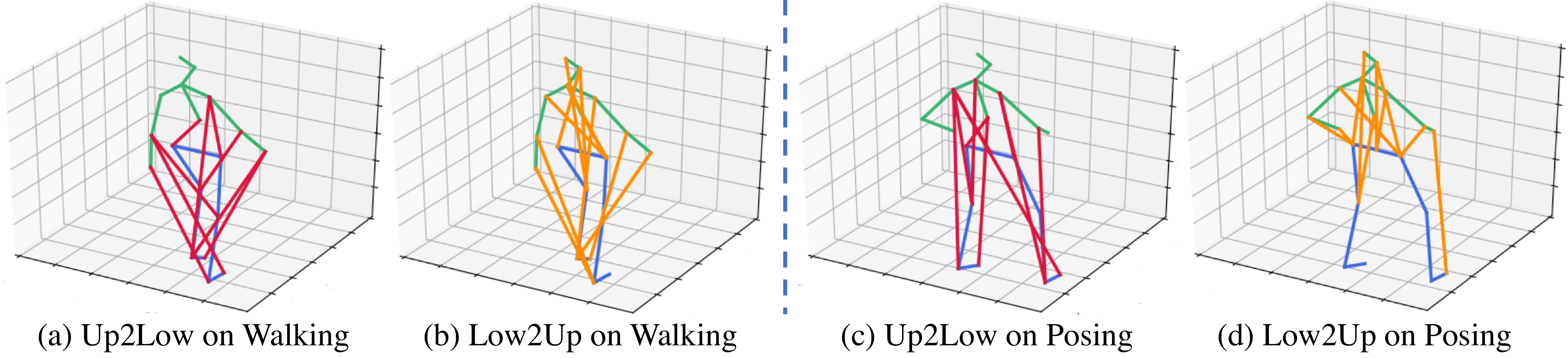}
    % \vspace{-3mm}
    \caption{\small Inferred directed bipartite cross-part graphs on posing and walking. We denote the upper-to-lower and the lower-to-upper graphs as `Up2Low' and `Low2Up'.}
    \label{fig:bodypart_inetraction}
    % \vspace{-5mm}
\end{figure}

To verify the inferred interactions between the upper and lower bodies, we visualize the learned bipartite graphs on different actions. We show the upper-to-lower and lower-to-upper graphs on `walking' and `posing', where we plot the edges whose weights are larger than $0.25$; see Fig.~\ref{fig:bodypart_inetraction}.
Different actions reflect different bipartite graphs: walking connects the contralateral hands and feet on both upper-to-lower and lower-to-upper graphs; posing connects the ipsilateral joints on the two body-parts, which delivers near-torso features to the body.
% \begin{figure}[t]
%   \begin{center}
%     \begin{tabular}{cccc}
%      \includegraphics[width=0.22\columnwidth]{figure/BipartGraph_Posing_Up2Low.png} & 
%      \includegraphics[width=0.22\columnwidth]{figure/BipartGraph_Posing_Low2Up.png} 
%      \includegraphics[width=0.22\columnwidth]{figure/BipartGraph_Walking_Up2Low.png} & 
%      \includegraphics[width=0.22\columnwidth]{figure/BipartGraph_Walking_Low2Up.png} 
%      \\
%      {{\small (a) Up2Low on posing.}} & 
%      {{\small (b) Low2Up on posing.}} &
%      {{\small (c) Up2Low on walking.}} & 
%      {{\small (d) Low2Up on walking.}}
%   \end{tabular}
% \end{center}
% %\vspace{-5mm}
% \caption{\small Inferred directed bipartite cross-part graphs on posing and walking. We denote the upper-to-lower graphs and the lower-to-upper graphs as `Up2Low' and `Low2Up' for simplification.}
% %\vspace{-3mm}
% \label{fig:PartInteraction}
% \end{figure}

\begin{table}[!h]
    \small
    \centering
    \caption{\small Comparison of running time, model sizes and MPJPE}
    % \vspace{-3mm}
    \resizebox{0.9\columnwidth}{!}{
    \begin{tabular}{|c|ccccc|c|}
      \hline
      ~ & DMGNN~\cite{Li_2020_CVPR} & Traj-GCN~\cite{Mao_2019_ICCV} & MSR-GCN~\cite{Dang_2021_ICCV} & STSGCN~\cite{Sofianos_2021_ICCV} & HisRep~\cite{Mao_2020_ECCV} & SPGSN (Ours) \\
      \hline
      RunTime (ms) & 33.13 & 26.35 & 42.36 & 17.6 & 31.57 & 30.07 \\
      ParaSize (M) & 4.82 & 2.56 & 6.30 & 0.04 & 3.18 & 5.66 \\
      FLOPs (M) & 2.82 & 0.49 & 3.89 & 1.35 & 3.08 & 1.77 \\
      MPJPE & 49.03 & 38.63 & 38.06 & 38.03 & 36.41 & {\bf 34.53} \\
      \hline
    \end{tabular}}
    \label{tab:Time_and_Size}
    % \vspace{-2mm}
\end{table}

\mypar{Efficiency Analysis}
To verify the applicability of SPGSN, we compare SPGSN to existing methods in terms of the running times, parameter numbers, FLOPs and prediction results in short-term prediction on H3.6M (Table~\ref{tab:Time_and_Size}).
SPGSN has the lowest MPJPE and efficient running based on the parallel computation. SPGSN also has the acceptable model size.

\section{Conclusion}
\label{sec:One_conclu}
%\vspace{-2mm}
We propose a novel skeleton graph scattering network for human motion prediction, which contains cascaded multi-part graph scattering blocks (MPGSBs) to capture fine representation along both spatial and spectrum dimensions. Each MPGSB builds adaptive graph scattering on separated body-parts. In this way, the model carries large graph spectrum and considers the distinct part-based dynamics for precise motion prediction. Experiments reveal the effectiveness of our model for motion prediction on Human3.6M, CMU Mocap and 3DPW datasets.

~

\noindent\textbf{Acknowledgements:}
This work is supported by 
the National Key Research and Development Program of China (2020YFB1406801), 
the National Natural Science Foundation of China under Grant (62171276),  
111 plan (BP0719010), 
STCSM (18DZ2270700, 21511100900),
State Key Laboratory of UHD Video and Audio Production and Presentation.

% \mypar{Limitation and future work} 
% To model diverse movement patterns, this paper proposes spatial decomposition to parse the human bodies. However, currently, the spatial decomposition is achieved with the pre-defined upper-lower body decomposition, which can not cover all possible cases, such as only the left body moving but the right keeping static. A possible direction is to explore more flexible and adaptive body decomposition methods, which could simultaneously separate the most uncorrelated body-parts for more effective pattern learning. Also, the reasonable body-parts separation and graph learning help to abandon the `whole body' to reduce the information redundancy.

\clearpage
\bibliographystyle{splncs04}
\bibliography{egbib}

\clearpage

\section*{I. Some details of Model}
\textbf{About SPGSN architectures.}
The entire SPGSN is constructed by cascaded MPGSBs, where we note that all the MPGSBs use distinct parameters without weight-sharing. The non-shared MPGSBs flexibly extract multi-level features to enrich the representation, just like ResNet uses distinct residual blocks to learn the hierarchical features.

Since the MPGSBs are formed in cascade, the input of the second MPGSB is the output of the first MPGSB, ${\bf H}'$, integrating the features from spatial and spectrum domains. We use IDCT to transform the output features to the predicted sequences.

~

\noindent \textbf{About experimental scenarios.}
The coordinates of all the body-joints denote the relative positions of distinct body-joints to a central joint (e.g. chest). In this benchmark, all the studies target to predict the pure poses, regardless of the body positions in a global space.

\section*{II. Protocol Modification of STSGCN}
Published in ICCV 2021, STSGCN~\cite{Sofianos_2021_ICCV} is one of the practical algorithms of human motion prediction, which leverages both spatial and temporal graph convolution on the different network stages. According to the official codes\footnote{https://github.com/FraLuca/STSGCN}, however, the experimental protocols have two main differences from most previous works~\cite{Martinez_2017_CVPR,Li_2018_CVPR,Gui_2018_ECCV,guo2019human,Mao_2019_ICCV,Li_2020_CVPR,Mao_2020_ECCV}, leading to their reported results having very different means:

$\bullet$ First, to evaluate the prediction performance at each timestamp, STSGCN learns a specific and unique model; that is, STSGCN uses different pre-trained models to generate poses at different timestamps such as 80ms and 160ms.
However, most previous models use only one model to forecast the whole pose sequences directly and just calculate the errors at different timestamps based on the same model.
The difference is that previous models generate one sequence only once to achieve low errors at any timestamps, while STSGCN builds specific models towards specific prediction timestamps, and the models are supervised by the target sequences within the corresponding timestamps.

$\bullet$ Second, to present the MPJPE at each timestamp, previous methods directly calculate the instant errors at the corresponding timestamp. However, according to the STSGCN's test codes, at a certain timestamp, STSGCN presents the average MPJPE before this timestamp instead of the instant one; that is, the 1000ms-MPJPE reported by STSGCN is the average MPJPE across all the prediction frames from the 0ms to the 1000ms.

To achieve a fair comparison, we modify the STSGCN codes to directly generate the whole sequence using only one model and reported the instant MPJPE at each timestamp in our manuscript.

\section*{III. The Body Separation}
\begin{figure}[t]
  \begin{center}
    \begin{tabular}{ccc}
     \includegraphics[width=0.26\columnwidth]{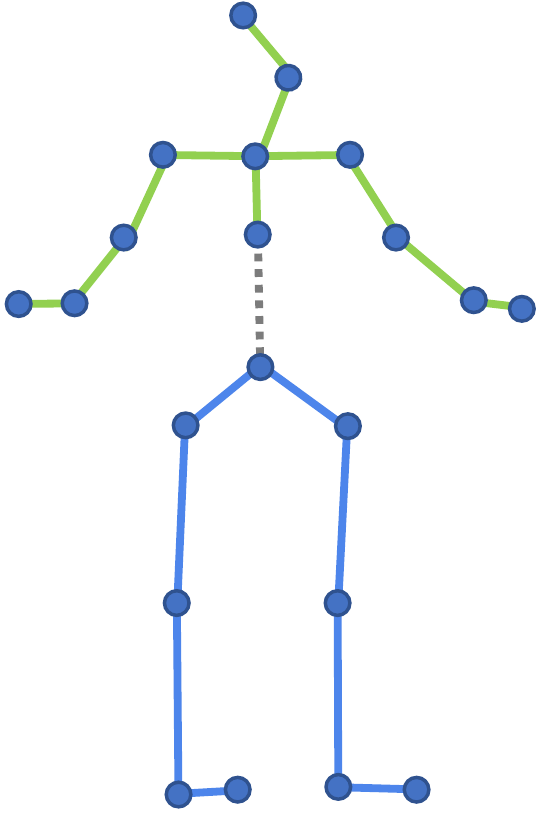} & 
     \includegraphics[width=0.26\columnwidth]{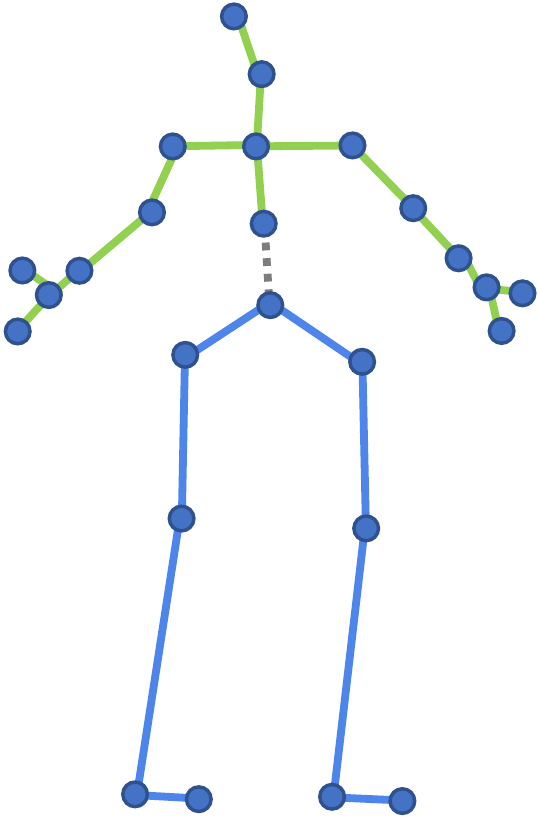} &
     \includegraphics[width=0.26\columnwidth]{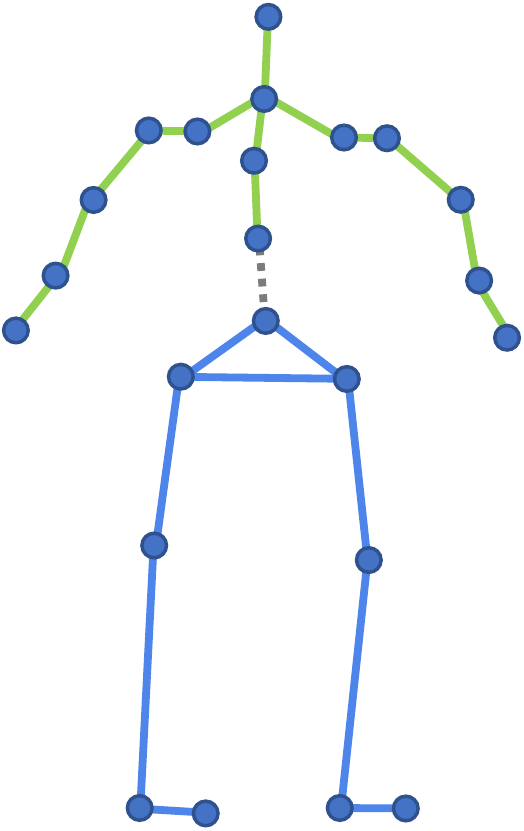} 
     \\
     {\small (a) Human 3.6M.} & 
     {\small (b) CMU Mocap.} & 
     {\small (c) 3DPW.} \\
  \end{tabular}
\end{center}
% \vspace{-5mm}
\caption{\small The body separation strategies on various datasets, where we color the bones on upper bodies in green and the bones on lower bodies in blue.}
\label{fig:PartSep}
\end{figure}

Here we illustrate the body separation strategies on the three datasets: Human 3.6M, CMU Mocap and 3DPW. We show the decomposed upper bodies and lower bodies in Fig.~\ref{fig:PartSep}.
We color the bones on the separated upper bodies in green, and the bones on the lower bodies in blue. Different datasets show different predefined body structures, thus we should use specific body separation strategies to obtain valid body-parts.

\section*{IV. Effects of the spectrum aggregation}
To study the proposed graph spectrum aggregation, we first compare our method to two common feature aggregation operations. First, directly averaging all the channels; second, summing up channels with an inner-product-based inter-channel self-attention. Table~\ref{tab:ablation_aggregation} presents the prediction results of SPGSN with various aggregation methods on H3.6M.
\begin{table}[t]
      \footnotesize  
      \centering
      \caption{\small SPGSNs with different spectrum aggregation methods.}
      %% \vspace{-3mm}
      \renewcommand{\arraystretch}{1.0}
      \resizebox{0.6\columnwidth}{!}{
      \begin{tabular}{|c|cccc|cc|}
          \hline
          ~ & \multicolumn{6}{c|}{Average MAE} \\
          \hline
          Model & 80 & 160 & 320 & 400 & 560 & 1000 \\
          \hline
          SPGSN (average) & 10.54 & 22.48 & 47.73 & 58.91 & 79.20 & 112.14 \\
          SPGSN (self-att.) & 10.82 & 22.79 & 47.65 & 59.13 & 79.61 & 114.08 \\
          SPGSN & 10.44 & 22.33 & 47.07 & 58.26 & 77.40 & 109.64 \\
          \hline
      \end{tabular}}
      %% \vspace{-2mm}
      \label{tab:ablation_aggregation}
\end{table}
We see that, the proposed graph spectrum aggregation outperforms the two variants. Note that, the self-attention mixes up various channels before aggregation but loses the spectrum diversity, leading to larger errors.

\begin{figure}[!t]
    \centering
    \includegraphics[width=0.9\columnwidth]{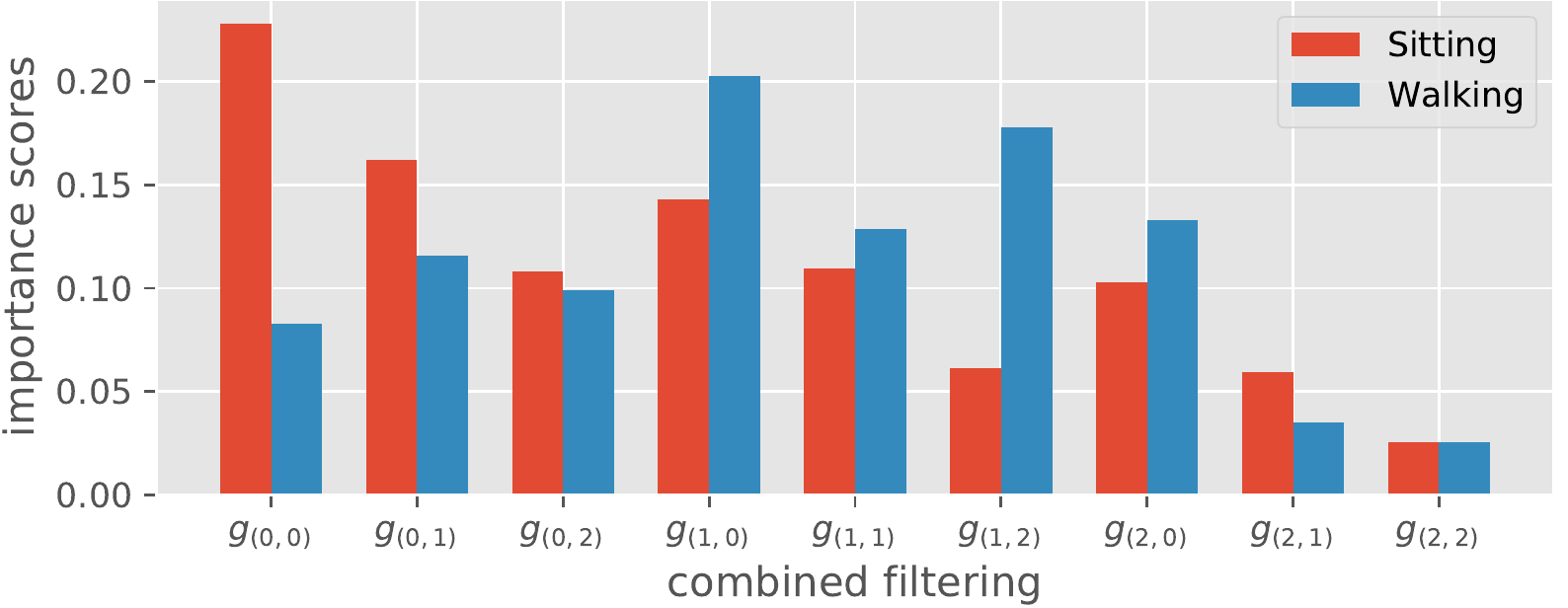}
    %% \vspace{-4mm}
    \caption{\small Spectrum importance scores on different actions.}
    \label{fig:AttentionPlot}
    %% \vspace{-5mm}
\end{figure}
Next, we visualize the learned spectrum importance of different actions. We show the importance scores calculated by the last MPGSB on actions `Sitting' and `Walking'; see Fig.~\ref{fig:AttentionPlot}, where the x-axis denotes the graph scattering branch with different filter combinations.
We see that, different actions lead to different importance distributions. For `Sitting', the poses show slow movements, thus the low-pass features dominate the spectrum to stabilize pattern learning; for `Walking', poses keep large movements, thus some high-frequency bands are emphasized.

\section*{V. Feature Responses of the Adaptive Graph Scattering Decomposition}
\begin{figure}
    \centering
    \includegraphics[width=0.9\columnwidth]{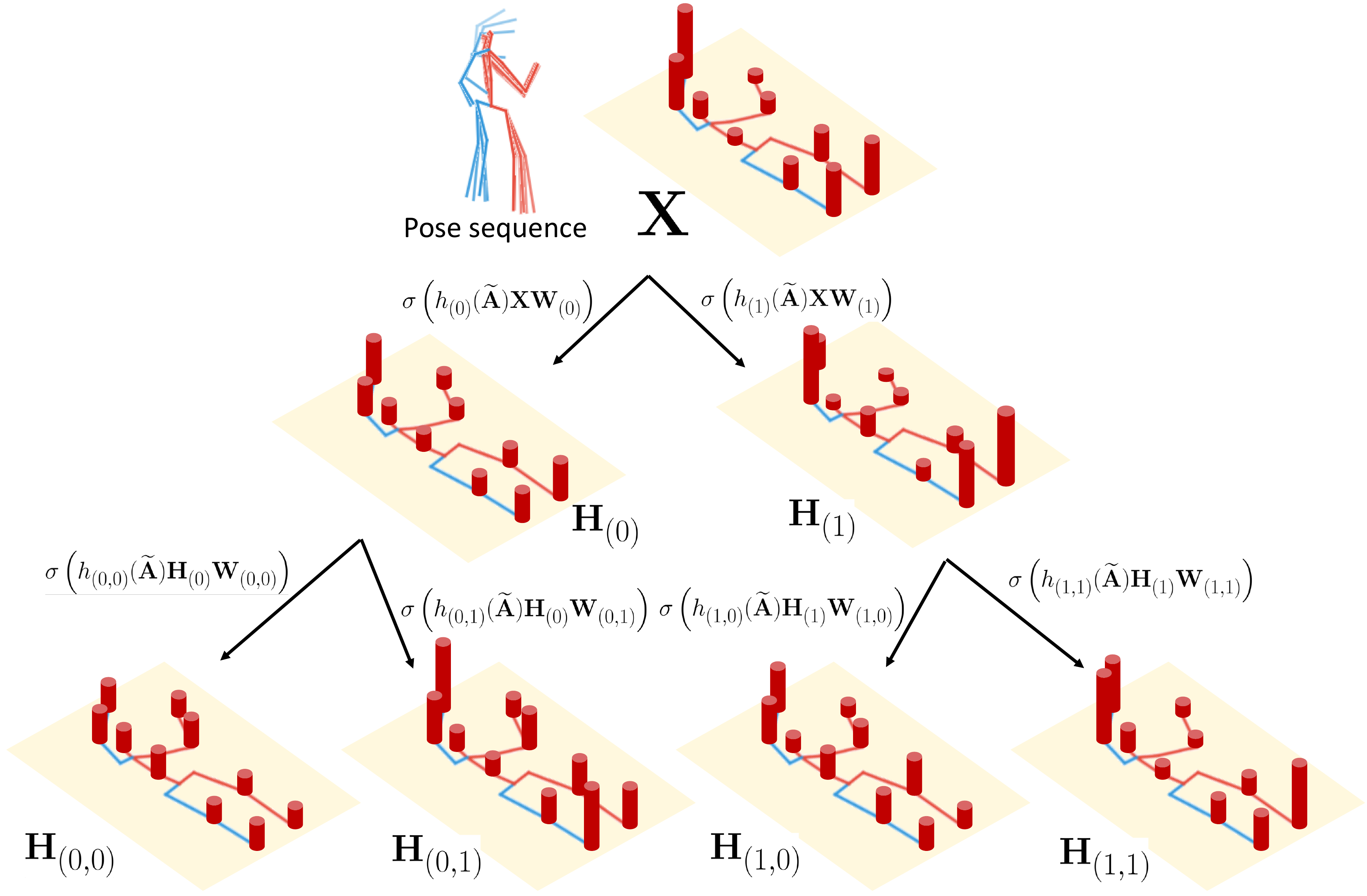}
    \caption{Feature responses after the proposed adaptive graph scattering decomposition of a pose sequence of the action 'directions'.}
    \label{fig:scattering_test}
\end{figure}
To study the effects of the proposed adaptive graph scattering decomposition and provide a more interpretable analysis, we consider showing the feature responses of the adaptive graph scattering decomposition given the input motion sequence. We select a sample of the action ’direction’, where the subject has a static body torso as well as performs hand raising and leg swaying. Since each body joint carries a DCT-formed multidimensional feature vector, we collect the second DCT coefficients of all the joints as a graph signal, on which we perform the graph scattering. The second DCT coefficients carry large values at the highly dynamic joints but carry small values at relatively fixed joints, according to empirical tests. Therefore, it is convenient to reflect on how much information about the dynamic joints is preserved. Moreover, the selected graph signal also shows the difference between nodes, which can reflect graph scattering extract the feature at various graph spectral bands, and better demonstrate the performance of the model.

The feature responses at each channel in each layer of the graph scattering decomposition are visualized in Fig.~\ref{fig:scattering_test}. We crop a sub-tree of graph scattering decomposition from the pre-trained model, where each non-leaf channel has only two child branches instead of three child branches in the original model, for clearer presentation. We use two filters at each non-leaf channel; that is $h_{(0)}(\widetilde{\bf A})$, a basic graph convolution, and $h_{(1)}(\widetilde{\bf A})$, an adaptive graph wavelet; see Eq. (2) in our manuscript. We also apply the feature transform parameter ${\bf W}$ and nonlinear activation $\sigma(\cdot)$ at each channel. In Fig.~\ref{fig:scattering_test}, the amplitude of the graph signals is plotted on the corresponding body-joints in the form of a histogram.

Fig.~\ref{fig:scattering_test} reveals that, given the motion data with large signal values at moving joints and small values at static joints, the graph convolution tends to average their features, close their difference and derive a smoother signal; especially in ${\bf H}_{(0,0)}$, the differences between the high-speed limbs and the fixed torso are hardly reflected. This means that always using deep graph convolution will not be able to accurately describe the characteristics of different body joints, and it is difficult to provide rich and detailed information for pattern recognition. As for graph wavelet filtering, the difference between joints could be effectively preserved; that is, in this sample, the graph wavelet filtering focuses on the high-frequency information to push joints farther and magnify their gaps. Note that, in the graph scattering, all the filtered channels covering both low-frequency and high-frequency bands are important, since the large spectrum contains significantly comprehensive information to promote pattern learning and motion prediction.

\section*{VI. Effects of Skip-connections across the Model}
Table~\ref{tab:ablation_skipconnection} compares SPGSN with and without the skip-connections on the H3.6M dataset. Skip-connections force to learn the feature displacements from observations to predictions, obtaining much smaller MPJPEs (especially in the short-term prediction) than the model without skip-connections that directly generates the poses.
\vspace{-3mm}
\begin{table}[b]
      \vspace{-3mm}
      \small 
      \centering
      \caption{\small Prediction with/without using skip-connections.}
      \renewcommand{\arraystretch}{1.0}
      \resizebox{0.6\columnwidth}{!}{
      \begin{tabular}{|c|cccc|cc|}
          \hline
          ~ & \multicolumn{6}{c|}{Average MPJPE} \\
          \hline
          millisecond & 80 & 160 & 320 & 400 & 560 & 1000 \\
          \hline
          No-skip & 23.19 & 31.81 & 54.15 & 64.05 & 81.90 & 115.22 \\
          Skip & 10.44 & 22.33 & 47.07 & 58.26 & 77.40 & 109.64 \\
          \hline
      \end{tabular}}
      \label{tab:ablation_skipconnection}
\end{table}

\section*{VII. More Quantitative Results}
Here we present the prediction results on all the contained actions in H3.6M and CMU Mocap for both short-term and long-term prediction. These results provide sufficient information for a detailed comparison of the algorithm development in future works.

First, we present the MPJPE of various models on H3.6M for short-term motion prediction, where the detailed results of any actions are shown in Table~\ref{tab:h36m_short_3D}.
\begin{table*}[t]
    \centering
    \small
    \caption{\small Prediction MPJPEs of various models for short-term motion prediction on H3.6M and the average MPJPEs across all the actions.}
    % \vspace{-2mm}
    \renewcommand{\arraystretch}{1.0}
    \resizebox{1\textwidth}{!}{
    \begin{tabular}{|c|cccc|cccc|cccc|cccc|}
        \hline
        Motion & \multicolumn{4}{c|}{Walking} & \multicolumn{4}{c|}{Eating} & \multicolumn{4}{c|}{Smoking} & \multicolumn{4}{c|}{Discussion} \\
        % \hline
        millisecond & 80&160&320&400 & 80&160&320&400 & 80&160&320&400 & 80&160&320&400 \\
        \hline
        Res-sup~\cite{Martinez_2017_CVPR} & 29.36 & 50.82 & 76.03 & 81.52 & 16.84 & 30.60 & 56.92 & 68.65 & 22.96 & 42.64 & 70.24 & 83.68 & 32.94 & 61.18 & 90.92 & 96.19 \\
        CSM~\cite{Li_2018_CVPR} & 21.70 & 43.56 & 66.29 & 75.48 & 14.50 & 26.13 & 47.47 & 55.63 & 19.42 & 37.70 & 62.49 & 68.55 & 26.35 & 53.41 & 79.12 & 83.01 \\
        SkelNet~\cite{guo2019human} & 20.49 & 34.36 & 59.64 & 68.76 & 11.80 & 22.38 & 39.88 & 48.11 & 11.33 & 23.71 & 45.30 & 52.85 & 21.79 & 40.24 & 65.93 & 77.91 \\
        DMGNN~\cite{Li_2020_CVPR} & 17.32 & 30.67 & 54.56 & 65.20 & 10.96 & 21.39 & 36.18 & 43.88 & 8.97 & 17.62 & 32.05 & 40.30 & 17.33 & 34.78 & 61.03 & 69.80 \\
        Traj-GCN~\cite{Mao_2019_ICCV} & 12.29 & 23.03 & 39.77 & 46.12 & 8.36 & 16.90 & 33.19 & 40.70 & 7.94 & 16.24 & 31.90 & 38.90 & 12.50 & 27.40 & 58.51 & 71.68 \\
        HisRep~\cite{Mao_2020_ECCV} & 10.53 & 19.96 & 34.88 & {42.05} & 7.39 & 15.53 & 31.26 & 38.58 & 7.17 & 14.54 & 28.83 & 35.67 & 10.89 & 25.19 & 56.15 & 69.30 \\
        MSR-GCN~\cite{Dang_2021_ICCV} & 12.16 & 22.65 & 38.64 & 45.24 & 8.39 & 17.05 & 33.03 & 40.43 & 8.02 & 16.27 & 31.32 & 38.15 & 11.98 & 26.76 & 57.08 & 69.74 \\
        STSGCN*~\cite{Sofianos_2021_ICCV} & 16.26 & 24.63 & 40.06 & 45.94 & 14.32 & 22.14 & 37.91 & 45.03 & 13.10 & 20.20 & 37.71 & 44.65 & 14.33 & 24.28 & {\bf 52.62} & {68.53} \\
        \hline
        SPGSN (1body) & {\bf 10.13} & 19.51 & 35.52 & 44.67 & 7.13 & 15.02 & 31.87 & 41.18 & 6.83 & 13.94 & 28.77 & 36.78 & 10.42 & 23.90 & 54.13 & 69.99 \\
        SPGSN & 10.14 & {\bf 19.39} & {\bf 34.80} & {\bf 41.47} & {\bf 7.07} & {\bf 14.85} & {\bf 30.48} & {\bf 37.91} & {\bf 6.72} & {\bf 13.79} & {\bf 27.97} & {\bf 34.61} & {\bf 10.37} & {\bf 23.79} & 53.61 & {\bf 67.12} \\
        \hline
        Motion & \multicolumn{4}{c|}{Directions} & \multicolumn{4}{c|}{Greeting} & \multicolumn{4}{c|}{Phoning} & \multicolumn{4}{c|}{Posing} \\
        millisecond & 80&160&320&400 & 80&160&320&400 & 80&160&320&400 & 80&160&320&400 \\
        \hline
        Res-sup~\cite{Martinez_2017_CVPR} & 35.36 & 57.27 & 76.30 & 87.67 & 34.46 & 63.36 & 124.60 & 142.50 & 37.96 & 69.32 & 115.00 & 126.73 & 36.10 & 69.12 & 130.46 & 157.08 \\
        CSM~\cite{Li_2018_CVPR} & 27.07 & 44.72 & 63.94 & 75.37 & 28.63 & 60.69 & 119.25 & 139.92 & 25.66 & 40.13 & 63.06 & 78.01 & 22.02 & 40.34 & 93.65 & 119.32 \\
        SkelNet~\cite{guo2019human} & 16.06 & 27.12 & 62.97 & 72.75 & 24.71 & 56.90 & 111.74 & 134.25 & 18.91 & 34.69 & 59.34 & 72.09 & 18.51 & 34.67 & 80.83 & 106.39 \\
        DMGNN~\cite{Li_2020_CVPR} & 13.14 & 24.62 & 64.68 & 81.86 & 23.30 & 50.32 & 107.30 & 132.10 & 12.47 & 25.77 & 48.08 & 58.29 & 15.27 & 29.27 & 71.54 & 96.65 \\
        Traj-GCN~\cite{Mao_2019_ICCV} & 8.97 & 19.87 & 43.35 & 53.74 & 18.65 & 38.68 & 77.74 & 93.39 & 10.24 & 21.02 & 42.54 & 52.30 & 13.66 & 29.89 & 66.62 & 84.05 \\
        HisRep~\cite{Mao_2020_ECCV} & 7.77 & 18.23 & 41.34 & 51.61 & 15.47 & 34.04 & 73.77 & 88.90 & 9.78 & 20.98 & 39.81 & 50.87 & 13.23 & 27.70 & 63.68 & 81.82 \\
        MSR-GCN~\cite{Dang_2021_ICCV} & 8.61 & 19.65 & 43.28 & 53.82 & 16.48 & 36.95 & 77.32 & 93.38 & 10.10 & 20.74 & 41.51 & 51.26 & 12.79 & 29.38 & 66.95 & 85.01 \\
        STSGCN*~\cite{Sofianos_2021_ICCV} & 14.24 & 24.27 & 44.24 & 53.21 & 15.02 & {\bf 30.70} & {\bf 67.11} & 87.63 & 14.88 & 21.40 & 46.55 & 52.03 & 15.01 & 25.69 & {\bf 58.38} & {\bf 73.08} \\
        \hline
        SPGSN (1body) & 7.38 & 17.48 & 40.54 & 53.09 & 15.16 & 33.61 & 71.89 & 88.74 & 8.78 & 18.50 & 39.85 & 51.53 & 10.92 & 25.46 & 61.38 & 78.87 \\
        SPGSN & {\bf 7.35} & {\bf 17.15} & {\bf 39.80} & {\bf 50.25} & {\bf 14.64} & 32.59 & 70.64 & {\bf 86.44} & {\bf 8.67} & {\bf 18.32} & {\bf 38.73} & {\bf 48.46} & {\bf 10.73} & {\bf 25.31} & 59.91 & 76.46 \\
        \hline
        Motion & \multicolumn{4}{c|}{Purchases} & \multicolumn{4}{c|}{Sitting} & \multicolumn{4}{c|}{Sitting Down} & \multicolumn{4}{c|}{Taking Photo} \\
        millisecond & 80&160&320&400 & 80&160&320&400 & 80&160&320&400 & 80&160&320&400 \\
        \hline
        Res-sup~\cite{Martinez_2017_CVPR} & 36.33 & 60.30 & 86.53 & 95.92 & 42.55 & 81.40 & 134.70 & 151.78 & 47.28 & 85.95 & 145.75 & 168.86 & 26.10 & 47.61 & 81.40 & 94.73 \\
        CSM~\cite{Li_2018_CVPR} & 25.69 & 47.85 & 82.49 & 93.90 & 22.25 & 34.67 & 58.72 & 75.80 & 23.67 & 51.76 & 102.93 & 119.47 & 20.29 & 38.92 & 61.14 & 77.40 \\
        SkelNet~\cite{guo2019human} & 21.04 & 40.59 & 79.97 & 88.66 & 15.55 & 28.70 & 49.35 & 62.87 & 17.64 & 38.88 & 85.30 & 101.71 & 15.74 & 32.83 & 48.62 & 63.90 \\
        DMGNN~\cite{Li_2020_CVPR} & 21.35 & 38.71 & 75.67 & 82.74 & 11.92 & 25.11 & 44.59 & {\bf 50.20} & 14.95 & 32.88 & 77.06 & 93.00 & 13.61 & 28.95 & 45.99 & 58.76 \\
        Traj-GCN~\cite{Mao_2019_ICCV} & 15.60 & 32.78 & 65.72 & 79.25 & 10.62 & 21.90 & 46.33 & 57.91 & 16.14 & 31.12 & 61.47 & 75.46 & 9.88 & 20.89 & 44.95 & 56.58 \\
        HisRep~\cite{Mao_2020_ECCV} & 14.63 & 32.81 & 65.18 & 78.27 & 10.21 & 20.36 & 43.68 & 53.62 & 15.54 & 29.97 & 59.31 & 72.25 & 9.09 & 20.10 & 44.60 & 55.72 \\
        MSR-GCN~\cite{Dang_2021_ICCV} & 14.75 & 32.39 & 66.13 & 79.64 & 10.53 & 21.99 & 46.26 & 57.80 & 16.10 & 31.63 & 62.45 & 76.84 & 9.89 & 21.01 & 44.56 & 56.30 \\
        STSGCN*~\cite{Sofianos_2021_ICCV} & 15.26 & {\bf 26.26} & 63.45 & {\bf 74.25} & 15.19 & 22.95 & 46.82 & 58.34 & 16.70 & 28.05 & {\bf 56.15} & 72.03 & 16.61 & 24.84 & 45.98 & 61.79 \\
        \hline
        SPGSN (1body) & 12.78 & 28.86 & 62.59 & 77.01 & {\bf 9.25} & 19.58 & 43.47 & 56.32 & 14.34 & 28.10 & 58.23 & 74.44 & {\bf 8.72} & 18.95 & 42.62 & 55.22 \\
        SPGSN & {\bf 12.75} & 28.58 & {\bf 61.01} & 74.38 & 9.28 & {\bf 19.40} & {\bf 42.25} & {53.56} & {\bf 14.18} & {\bf 27.72} & 56.75 & {\bf 70.74} & 8.79 & {\bf 18.90} & {\bf 41.49} & {\bf 52.66} \\
        \hline
        Motion & \multicolumn{4}{c|}{Waiting} & \multicolumn{4}{c|}{Walking Dog} & \multicolumn{4}{c|}{Walking Together} & \multicolumn{4}{c|}{Average} \\
        millisecond & 80&160&320&400 & 80&160&320&400 & 80&160&320&400 & 80&160&320&400 \\
        \hline
        Res-sup~\cite{Martinez_2017_CVPR} & 30.62 & 57.82 & 106.22 & 121.45 & 64.18 & 102.10 & 141.07 & 164.35 & 26.79 & 50.07 & 80.16 & 92.23 & 34.66 & 61.97 & 101.08 & 115.49 \\
        CSM~\cite{Li_2018_CVPR} & 19.14 & 33.11 & 69.72 & 95.21 & 58.67 & 97.36 & 129.74 & 158.57 & 22.60 & 38.51 & 71.13 & 84.37 & 25.17 & 45.92 & 78.08 & 93.33 \\
        SkelNet~\cite{guo2019human} & 16.31 & 29.90 & 63.86 & 84.59 & 54.61 & 93.23 & 124.12 & 155.79 & 19.01 & 32.40 & 63.35 & 73.18 & 20.23 & 38.04 & 69.35 & 84.25 \\
        DMGNN~\cite{Li_2020_CVPR} & 12.20 & 24.17 & 59.62 & 77.54 & 47.09 & 93.33 & 160.13 & 171.20 & 14.34 & 26.67 & 50.08 & 63.22 & 16.95 & 33.62 & 65.90 & 79.65 \\
        Traj-GCN~\cite{Mao_2019_ICCV} & 11.43 & 23.99 & 50.06 & 61.48 & 23.39 & 46.17 & 83.47 & 95.96 & 10.47 & 21.04 & 38.47 & 45.19 & 12.68 & 26.06 & 52.27 & 63.51 \\
        HisRep~\cite{Mao_2020_ECCV} & 10.58 & 23.75 & 49.30 & 60.26 & 21.77 & 43.38 & 78.53 & 90.21 & 9.88 & 19.51 & 35.91 & 42.60 & 11.60 & 24.40 & 49.75 & 60.78 \\
        MSR-GCN~\cite{Dang_2021_ICCV} & 10.68 & 23.06 & 48.25 & 59.23 & 20.65 & 42.88 & 80.35 & 93.31 & 10.56 & 20.92 & 37.40 & 43.85 & 12.11 & 25.56 & 51.64 & 62.93 \\
        STSGCN*~\cite{Sofianos_2021_ICCV} & 16.30 & 27.33 & 48.12 & 59.79 & {\bf 16.48} & {37.63} & {\bf 70.60} & {86.33} & 11.38 & 22.39 & 39.90 & 47.48 & 15.34 & 25.52 & 50.64 & 60.61 \\
        \hline
        SPGSN (1body) & 9.24 & 20.02 & 43.80 & 56.80 & 18.31 & 38.12 & 73.63 & 86.74 & {\bf 8.91} & 18.46 & 34.88 & 42.98 & 10.55 & 22.63 & 48.21 & 60.96 \\
        SPGSN & {\bf 9.21} & {\bf 19.79} & {\bf 43.10} & {\bf 54.14} & 17.83 & {\bf 37.15} & 71.74 & {\bf 84.91} & 8.94 & {\bf 18.19} & {\bf 33.84} & {\bf 40.88} & {\bf 10.44} & {\bf 22.33} & {\bf 47.07} & {\bf 58.26} \\
        \hline
    \end{tabular}}
    % \vspace{-2mm}
    \label{tab:h36m_short_3D}
\end{table*}
We see that, the proposed SPGSN with spatial and spectrum feature decomposition could effectively outperform baselines at most actions, as well as achieve the best results in terms of the average prediction errors. Moreover, using the separated body parts could improve the prediction compared to the model variant using only the whole body.

Also, we show the prediction MPJPEs on H3.6M for long-term motion prediction, which are presented in Table~\ref{tab:h36m_long_MPJPE}.
\begin{table*}[t]
    \centering
    \caption{\small Prediction MPJPEs of methods for long-term prediction on H3.6M and the average MPJPEs across all the actions.}
    % \vspace{-2mm}
    \renewcommand{\arraystretch}{1.0}
    \footnotesize
    \resizebox{1\textwidth}{!}{
    \begin{tabular}{|c|cc|cc|cc|cc|cc|cc|cc|cc|}
        \hline
        Motion & 
        \multicolumn{2}{c|}{Walking}& \multicolumn{2}{c|}{Eating}&
        \multicolumn{2}{c|}{Smoking}& \multicolumn{2}{c|}{Discussion} &
        \multicolumn{2}{c|}{Directions} & \multicolumn{2}{c|}{Greeting} &
        \multicolumn{2}{c|}{Phoning} & \multicolumn{2}{c|}{Posing} \\
        millisecond & 560 & 1k & 560 & 1k & 560 & 1k & 560 & 1k & 560 & 1k & 560 & 1k & 560 & 1k & 560 & 1k \\
        \hline
        Res-sup~\cite{Martinez_2017_CVPR} & 81.73 & 100.68 & 79.87 & 100.20 & 94.83 & 137.44 & 121.30 & 161.70 & 110.05 & 152.48 & 156.32 & 184.29 & 143.92 & 186.79 & 165.41 & 236.79 \\
        CSM~\cite{Li_2018_CVPR} & 78.04 & 94.58 & 72.14 & 96.87 & 66.61 & 89.80 & 108.20 & 142.13 & 97.80 & 132.82 & 151.50 & 175.37 & 83.46 & 127.55 & 137.72 & 210.90 \\
        SkelNet~\cite{guo2019human} & 73.58 & 91.84 & 63.58 & 90.88 & 58.96 & 80.53 & 98.28 & 135.68 & 93.77 & 124.89 & 148.38 & 168.06 & 75.42 & 113.34 & 131.90 & 196.21 \\
        Traj-GCN~\cite{Mao_2019_ICCV} & 54.05 & 59.75 & 53.39 & 77.75 & 50.74 & 72.62 & 91.61 & 121.53 & 71.01 & 101.79 & 113.87 & 145.19 & 69.55 & 104.19 & 114.52 & 171.10 \\ 
        DMGNN~\cite{Li_2020_CVPR} & 71.36 & 85.82 & 58.11 & 86.66 & 50.85 & 72.15 & {\bf 81.90} & {\bf 106.32} & 102.06 & 135.75 & 144.51 & 170.54 & 71.33 & 108.37 & 125.45 & 188.18 \\ 
        MSR-GCN~\cite{Dang_2021_ICCV} & 52.72 & 63.05 & 52.54 & 77.11 & 49.45 & 71.64 & 88.59 & 117.59 & 71.18 & 100.59 & 116.24 & 147.23 & 68.28 & 104.36 & 116.26 & 174.33 \\
        STSGCN*~\cite{Sofianos_2021_ICCV} & 57.64 & 66.74 & 58.46 & 75.08 & 55.55 & 74.13 & 84.20 & 107.74 & 75.61 & 109.89 & {\bf 79.32} & {\bf 103.75} & 79.19 & 109.88 & 80.82 & 107.58 \\
        \hline
        SPGSN & {\bf 46.89} & {\bf 53.59} & {\bf 49.76} & {\bf 73.39} & {\bf 46.68} & {\bf 68.62} & 89.68 & 118.55 & {\bf 70.05} & {\bf 100.52} & 110.98 & 143.21 & {\bf 66.70} & {\bf 102.52} & 110.34 & 165.39 \\ 
        \hline
        Motion & 
        \multicolumn{2}{c|}{Purchases}& \multicolumn{2}{c|}{Sitting}&
        \multicolumn{2}{c|}{SittingDown}& \multicolumn{2}{c|}{TakingPhoto} &
        \multicolumn{2}{c|}{Waiting} & \multicolumn{2}{c|}{WalkingDog} &
        \multicolumn{2}{c|}{WalkingToge} & \multicolumn{2}{c|}{Average} \\
        millisecond & 560 & 1k & 560 & 1k & 560 & 1k & 560 & 1k & 560 & 1k & 560 & 1k & 560 & 1k & 560 & 1k \\
        \hline
        Res-sup~\cite{Martinez_2017_CVPR} & 119.36 & 176.92 & 166.20 & 185.16 & 197.09 & 223.58 & 107.03 & 162.38 & 126.70 & 153.14 & 173.61 & 202.31 & 94.51 & 110.48 & 129.19 & 164.96 \\
        CSM~\cite{Li_2018_CVPR} & 113.44 & 167.61 & 98.04 & 134.70 & 148.87 & 196.75 & 94.75 & 144.52 & 106.03 & 125.60 & 168.71 & 183.42 & 93.90 & 106.60 & 107.94 & 141.95 \\
        SkelNet~\cite{guo2019human} & 109.51 & 155.72 & 84.76 & 127.11 & 125.89 & 184.24 & 86.43 & 130.41 & 90.49 & 112.02 & 166.65 & 176.79 & 79.07 & 99.25 & 99.11 & 132.46 \\
        Traj-GCN~\cite{Mao_2019_ICCV} & 99.24 & 137.28 & 77.63 & 118.36 & 100.91 & 157.32 & 78.73 & 120.06 & 79.08 & 103.83 & 138.24 & 150.63 & 51.67 & 61.10 & 81.07 & 113.01 \\ 
        DMGNN~\cite{Li_2020_CVPR} & 104.86 & 146.09 & 75.51 & {\bf 115.44} & 118.04 & 174.05 & 78.38 & 123.65 & 85.54 & 113.68 & 183.20 & 210.17 & 70.46 & 86.93 & 93.57 & 127.62 \\ 
        MSR-GCN~\cite{Dang_2021_ICCV} & 101.63 & 139.15 & 78.19 & 120.02 & 102.83 & 155.45 & 77.94 & 121.87 & 76.33 & 106.25 & 111.87 & 148.21 & 52.93 & 65.91 & 81.13 & 114.18 \\
        STSGCN*~\cite{Sofianos_2021_ICCV} & 87.10 & 119.26 & 82.32 & 119.83 & 92.60 & 129.67 & 87.70 & 119.79 & 86.41 & 118.04 & 86.79 & 118.33 & 75.33 & 95.83 & 80.66 & 113.33 \\
        \hline
        SPGSN & 96.53 & 133.88 & {\bf 75.00} & 116.24 & 98.94 & 149.88 & {\bf 75.58} & {\bf 118.22} & {\bf 73.50} & {\bf 103.62} & 102.37 & 137.96 & 49.84 & 60.86 & {\bf 77.40} & {\bf 109.64} \\ 
        \hline
    \end{tabular}}
    % \vspace{-2mm}
    \label{tab:h36m_long_MPJPE}
\end{table*}
We see that, SPGSN obtains the effective performance in long-term motion prediction, since SPGSN shows lower MPJPEs on most actions as well as the lowest average MPJPE over all actions.

Finally, we present the MPJPEs of various methods on all the actions of CMU Mocap; see Table~\ref{tab:pred_cmu}.
\begin{table*}[t]
    \centering
    \caption{\small Prediction MPJPEs of different methods on the 8 actions of CMU Mocap for both short-term and long-term motion prediction. We also present the average prediction results across all the actions.}
    % \vspace{-2mm}
    \footnotesize
    \renewcommand{\arraystretch}{1.0}
    \resizebox{1\textwidth}{!}{
        \begin{tabular}{|c|ccccc|ccccc|ccccc|ccccc|}
        \hline
        Motion & \multicolumn{5}{c|}{Basketball} & \multicolumn{5}{c|}{Basketball Signal} & \multicolumn{5}{c|}{Directing Traffic} & \multicolumn{5}{c|}{Jumping} \\ 
        % \hline
        millisecond & 80 & 160 & 320 & 400 & 1000 & 80 & 160 & 320 & 400 & 1000 & 80 & 160 & 320 & 400 & 1000 & 80 & 160 & 320 & 400 & 1000 \\ \hline
        Res-sup.~\cite{Martinez_2017_CVPR} & 15.45 & 26.88 & 43.51 & 49.23 & 88.73 & 20.17 & 32.98 & 42.75 & 44.65 & 60.57 & 20.52 & 40.58 & 75.38 & 90.36 & 153.12 & 26.85 & 48.07 & 93.50 & 108.90 & 162.84  \\
        DMGNN~\cite{Li_2020_CVPR} & 15.57 & 28.72 & 59.01 & 73.05 & 138.62 & 5.03 & 9.28 & 20.21 & 26.23 & 52.04 & 10.21 & 20.90 & 41.55 & 52.28 & 111.23 & 31.97 & 54.32 & 96.66 & 119.92 & 224.63 \\
        Traj-GCN~\cite{Mao_2019_ICCV} & 11.68 & 21.26 & 40.99 & 50.78 & 97.99 & 3.33 & 6.25 & 13.58 & 17.98 & 54.00 & 6.92 & 13.69 & 30.30 & 39.97 & 114.16 & 17.18 & 32.37 & 60.12 & 72.55 & 127.41 \\
        MST-GCN~\cite{Dang_2021_ICCV} & 10.28 & 18.94 & {\bf 37.68} & {\bf 47.03} & {\bf 86.96} & 3.03 & 5.68 & 12.35 & 16.26 & 47.91 & 5.92 & 12.09 & 28.36 & 38.04 & 111.04 & 14.99 & 28.66 & {\bf 55.86} & {\bf 69.05} & {\bf 124.79} \\
        STSGCN*~\cite{Sofianos_2021_ICCV} & 12.56 & 23.04 & 41.92 & 50.33 & 94.17 & 4.72 & 6.69 & 14.53 & 17.88 & 49.52 & 6.41 & 12.38 & 29.05 & 38.86 & 109.42 & 17.52 & 31.48 & 58.74 & 72.06 & 127.40 \\
        \hline
        SPGSN & {\bf 10.24} & {\bf 18.54} & 38.22 & 48.68 & 89.58 & {\bf 2.91} & {\bf 5.25} & {\bf 11.31} & {\bf 15.01} & {\bf 47.31} & {\bf 5.52} & {\bf 11.16} & {\bf 25.48} & {\bf 37.06} & {\bf 108.14} & {\bf 14.93} & {\bf 28.16} & 56.72 & 71.16 & 125.20 \\
        \hline
        %\cline{1-21}
        Motion & \multicolumn{5}{c|}{Running} & \multicolumn{5}{c|}{Soccer} & \multicolumn{5}{c|}{Walking} & \multicolumn{5}{c|}{Washing Window} \\ 
        % \cline{1-21}
        millisecond & 80 & 160 & 320 & 400 & 1000 & 80 & 160 & 320 & 400 & 1000 & 80 & 160 & 320 & 400 & 1000 & 80 & 160 & 320 & 400 & 1000 \\ 
        \cline{1-21}
        Res-sup.~\cite{Martinez_2017_CVPR} & 25.76 & 48.91 & 88.19 & 100.80 & 158.19 & 17.75 & 31.30 & 52.55 & 61.40 & 107.37 & 44.35 & 76.66 & 126.83 & 151.43 & 194.33 & 22.84 & 44.71 & 86.78 & 104.68 & 202.73 \\
        DMGNN~\cite{Li_2020_CVPR} & 17.42 & 26.82 & 38.27 & 40.08 & {\bf 46.40} & 14.86 & 25.29 & 52.21 & 65.42 & 111.90 & 9.57 & 15.53 & 26.03 & 30.37 & 67.01 & 7.93 & 14.68 & 33.34 & 44.24 & 82.84 \\
        Traj-GCN~\cite{Mao_2019_ICCV} & 14.53 & 24.20 & 37.44 & 41.10 & 51.73 & 13.33 & 24.00 & 43.77 & 53.20 & 108.26 & 6.62 & 10.74 & 17.40 & 20.35 & {\bf 34.41} & 5.96 & 11.62 & 24.77 & 31.63 & 66.85 \\
        MST-GCN~\cite{Dang_2021_ICCV} & 12.84 & 20.42 & 30.58 & 34.42 & 48.03 & 10.92 & 19.50 & 37.05 & 46.38 & {\bf 99.32} & {\bf 6.31} & 10.30 & 17.64 & 21.12 & 39.70 & 5.49 & 11.07 & 25.05 & 32.51 & 71.30 \\
        STSGCN*~\cite{Sofianos_2021_ICCV} & 16.70 & 27.58 & 36.15 & 36.42 & 55.34 & 13.49 & 25.24 & 39.87 & 51.58 & 109.63 & 7.18 & 10.99 & 17.84 & 22.61 & 44.12 & 6.79 & 12.10 & 24.92 & 36.66 & 69.48 \\
        \cline{1-21}
        SPGSN & {\bf 10.75} & {\bf 16.67} & {\bf 26.07} & {\bf 30.08} & 52.92 & {\bf 10.86} & {\bf 18.99} & {\bf 35.05} & {\bf 45.16} & 99.51 & 6.32 & {\bf 10.21} & {\bf 16.34} & {\bf 20.19} & 34.83 & {\bf 4.86} & {\bf 9.44} & {\bf 21.50} & {\bf 28.37} & {\bf 65.08} \\ 
        \cline{1-21}
        Motion & \multicolumn{5}{c|}{Average} \\ 
        millisecond & 80 & 160 & 320 & 400 & 1000 \\
        \cline{1-6}
        Res-sup.~\cite{Martinez_2017_CVPR} & 24.21 & 43.75 & 76.19 & 88.93 & 139.00 \\
        DMGNN~\cite{Li_2020_CVPR} & 14.07 & 24.44 & 45.90 & 55.45 & 104.33 \\
        Traj-GCN~\cite{Mao_2019_ICCV} & 9.94 & 18.02 & 33.55 & 40.95 & 81.85 \\
        MST-GCN~\cite{Dang_2021_ICCV} & 8.72 & 15.83 & 30.57 & 38.10 & 79.01 \\
        STSGCN*~\cite{Sofianos_2021_ICCV} & 10.80 & 18.19 & 31.18 & 41.05 & 81.76 \\
        \cline{1-6}
        SPGSN & {\bf 8.30} & {\bf 14.80} & {\bf 28.64} & {\bf 36.96} & {\bf 77.82} \\
        \cline{1-6}
    \end{tabular}}
    % \vspace{-5mm}
    \label{tab:pred_cmu}
\end{table*}
The experiment results also verify the effectiveness of the proposed SPGSN.

\section*{VIII. More Qualitative Results}
\begin{figure}[t]
    \centering
    \includegraphics[width=1\columnwidth]{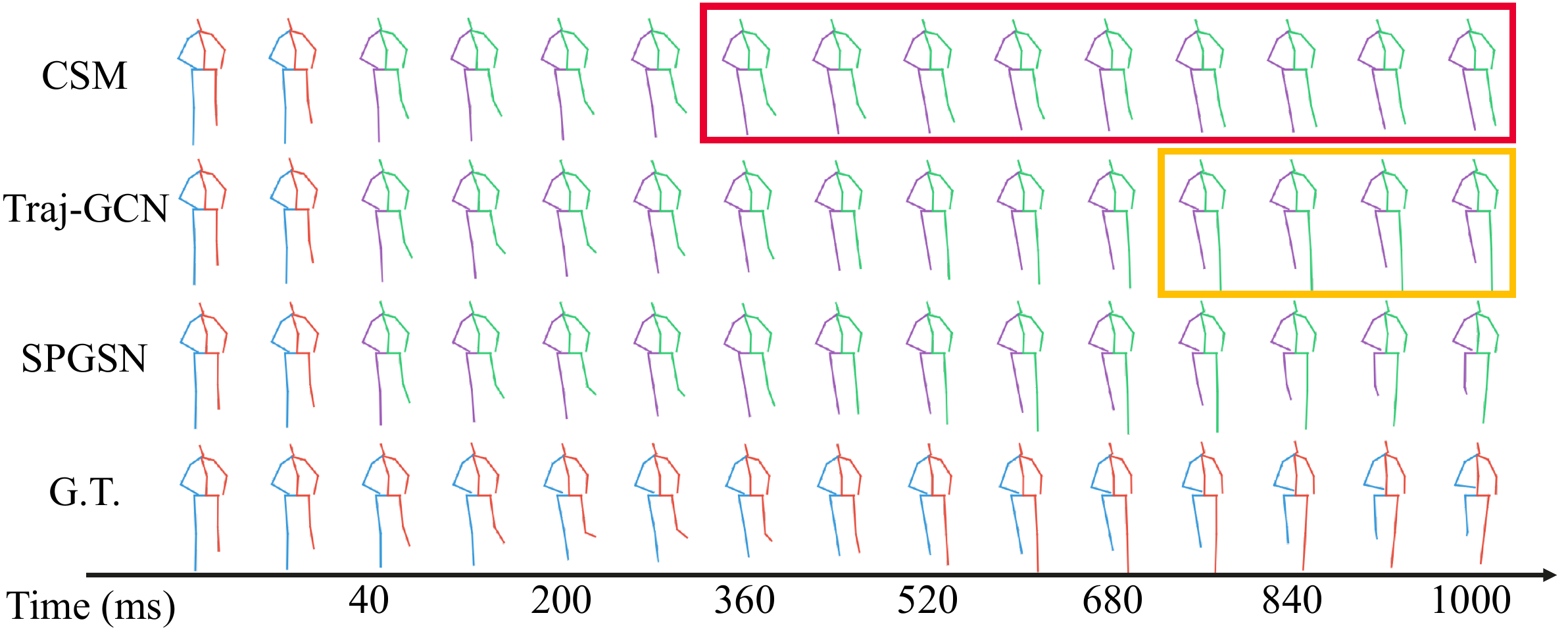}
    % \vspace{-2mm}
    \caption{\small The generated motion sequence of `walking dog' in H3.6M}
    \label{fig:prediction_sample_walkingdog}
    % \vspace{-5mm}
\end{figure}

Besides the Figure 5 in our manuscript, here we illustrate the predicted samples on the other action: walking dog, where we compare the proposed SPGSN to previous CSM~\cite{Li_2018_CVPR} and Traj-GCN~\cite{Mao_2019_ICCV}; see Figure~\ref{fig:prediction_sample_walkingdog}.
We see that, compared to the baselines, SPGSN generates more precise and reasonable future poses that are close to the ground truth in both short-term and long-term. For CSM, the predicted motion shows large errors after the $360$th ms; that is, the poses collapse to static where the right feet (green) are always lifted; see the red box. As for Traj-GCN, the generated poses keep static to lean to the left after the $720$th ms; see the orange box.

\end{document}